\documentclass[xcolor=dvipsnames]{scrarticle}
\usepackage[utf8]{inputenc}
\usepackage[a4paper, left=2cm, top=2cm, right=2.5cm, bottom=2cm]{geometry}
\usepackage{tikz}
\usepackage{natbib}
\usepackage{amssymb}
\usepackage{amsmath}
\usepackage{tabularx, ragged2e}
\usepackage{subcaption}
\usepackage{hyperref}
\usepackage[colorinlistoftodos, textsize=tiny]{todonotes}
\usepackage[croatian, english]{babel}
\usepackage{array}
\usepackage{graphicx}
\newcolumntype{Z}{>{\raggedright\let\newline\\\arraybackslash\hspace{0pt}}X}
\usepackage{eurosym}
\DeclareUnicodeCharacter{20AC}{\euro}
\usepackage[section]{placeins}
\usepackage{booktabs}
\usepackage{rotating}
\usepackage{tipa}
\usepackage{bm}
\usepackage{multirow}

\allowdisplaybreaks

\title{Frequency effects in Linear Discriminative Learning}
\author{Maria Heitmeier$^{1,2}$, Yu-Ying Chuang$^1$, Seth D. Axen$^2$, and R. Harald Baayen$^{1,2}$}

\date{$^1$Quantitative Linguistics, University of Tübingen\\ $^2$Cluster of Excellence Machine Learning: New Perspectives for Science, University of Tübingen}

\begin{document}
\maketitle

\begin{abstract}

Word frequency is a strong predictor in most lexical processing tasks \citep{brysbaert2011word}. Thus, any model of word recognition needs to account for how word frequency effects arise. The Discriminative Lexicon Model \citep[DLM;][]{baayen2018inflectional, baayen2019discriminative} models lexical processing with mappings between words' forms and their meanings. Comprehension and production are modelled via linear mappings between the two domains. So far, the mappings within the model can either be obtained incrementally via error-driven learning, a computationally expensive process able to capture frequency effects, or in an efficient, but frequency-agnostic solution modelling the theoretical endstate of learning (EL) where all words are learned optimally. In the present study we show how an efficient, yet frequency-informed mapping between form and meaning can be obtained (Frequency-informed learning; FIL). 
We find that FIL well approximates an incremental solution while being computationally much cheaper. FIL shows a relatively low type- and high token-accuracy, demonstrating that the model is able to process most word tokens encountered by speakers in daily life correctly. 
We use FIL to model reaction times in the Dutch Lexicon Project \citep{keuleers2010practice} by means of a Gaussian Location Scale Model and find that FIL predicts well the S-shaped relationship between frequency and the mean of reaction times but underestimates the variance of reaction times for low frequency words. FIL is also better able to account for priming effects in an auditory lexical decision task in Mandarin Chinese \citep{lee2007does}, compared to EL.  Finally, we used ordered data from CHILDES \citep{brown1973first, demuth2006word} to compare mappings obtained with FIL and incremental learning. We show that the mappings are highly correlated, but that with FIL some nuances based on word ordering effects are lost. Our results show how frequency effects in a learning model can be simulated efficiently, and raise questions about how to best account for low-frequency words in cognitive models.

\end{abstract}

\section{Introduction}

Word frequency effects are ubiquitous in psycholinguistic research. In fact, word frequency (i.e. the number of times a word occurs in some corpus) is one of the most important predictors in a range of psycholinguistic experimental paradigms \citep{brysbaert2011word}. In the lexical decision task, where participants are asked to decide whether a presented letter string is a word or not, frequency explains by far the most variance in reaction times, compared to other measures such as neighbourhood density or word length \citep[e.g.][]{baayen2010demythologizing, brysbaert2011word}: higher frequency words elicit faster reactions \citep[e.g.][]{rubenstein1970homographic, balota2004visual}. In word naming, another popular experimental paradigm in psycholinguistics where participants have to read aloud presented words, word frequency is less important, but still has a reliable effect: higher frequency words are named faster \citep[e.g.][]{balota2004visual}. Even though the effect of frequency has long been known and studied, to this day new studies are published confirming the effect in ever larger datasets across different languages \citep[e.g.][]{balota2004visual, keuleers2012british, keuleers2010practice, brysbaert2016impact, ferrand2010french}. Some studies have also proposed new frequency counts, explaining for example individual differences when it comes to the frequency effect \citep[e.g.][]{brysbaert2018word, kuperman2013reassessing}.

A central challenge for models of word recognition is therefore to explain word frequency effects. This challenge has been met in various different ways by influential models of word recognition. One of the earliest ideas proposed that words are stored in list-like structures ordered by frequency, such that the most frequent words are found earlier than lower frequency words \citep[e.g.][]{rubenstein1971homographic}. This idea was developed into the ``Search Model of Lexical Access'' by \citet{forster1976accessing, forster1979levels}. In this model there are ``peripheral access files'' in which words are stored according to their frequency of occurrence. These files hold ``access keys'' to where information about each word is stored in a main file. Words in the access files are grouped into bins based on form characteristics. Thus, the model aims to explain both word frequency effects (words are ordered according to their frequency in the peripheral access files) as well as neighbourhood effects (words with similar form are stored in the same bin). Later iterations of the model also suggested a hybrid model between serial and parallel search, where each bin is searched serially, but all bins are searched in parallel \citep{forster1994computational}.

The Logogen model \citep{morton1969interaction, morton1979facilitation, morton1979word} fully doubled down on the idea of a parallel search. A logogen can be seen as a detector for a set of input features. Every time one of its associated input features is encountered, the logogen's counter is increased. If the counter surpasses a threshold, it elicits a response. Each word/morpheme in a speaker's lexicon is assumed to correspond to a logogen. Additionally, there are logogens for lower level visual and auditory input features such as letters or phonemes whose outputs in turn serve as inputs to the word logogens. The Logogen model accounts for frequency effects by assuming that logogens corresponding to words with higher frequency have a lower response threshold than those corresponding to lower frequency words. After a logogen elicits a response, the threshold is lowered, and it takes a long time for the threshold to increase again. This explains trial-to-trial effects because just activated words will be activated faster in subsequent trials, but it also explains the long-term effect of word frequency because words occurring regularly will always have a lower threshold for eliciting a response \citep{morton1969interaction}.

The interactive activation model \citep{mcclelland1981interactive, rumelhart1982interactive} is in many ways a successor of the Logogen model. It proposes three levels of representations: one for letter features, one for letters and one for words. There are excitatory and inhibitory connections from letter features to letters and from letters to words. Additionally, there are excitatory and inhibitory connections from words to letters. Finally, there are both excitatory and inhibitory connections within each representational level (note though that the feature-feature inhibition was set to zero in the original implementation). The interactive activation model was originally proposed to explain the word superiority effect, i.e. the finding that letters are identified faster within a word than within a random letter string \citep[e.g.][]{reicher1969perceptual}. However, the model also proposes an account of frequency effects: the resting activation level of the word units are set depending on word frequency, such that high frequency words have a higher resting activation than low frequency words. In this way, \citet{jacobs1994models} were able to show that the interactive activation model shows the same effect of frequency on reaction times in various lexical decision experiments as human participants.

The interactive activation model was followed by the triangle model \citep{seidenberg1989distributed, harm2004computing}, which consists of distributed representations for orthography, phonology and semantics, each connected via a set of hidden units. In contrast to the interactive activation model, the weights between layers in this network are not set by the modeller but learned using backpropagation \citep{rumelhart1986backprop}. Thus, the error between the model's prediction and the actual target is reduced each time an input-target pair is encountered. For example, to model reading, the triangle model takes as input a word's orthography and predicts its phonology. Then, the error between the predicted phonology and the correct phonology of the word is computed, and the model's weights are updated such that the next time the same phonology is to be produced from the same orthography, it will be more accurate. The more often a word is presented to the model, the more accurate its predicted phonology becomes. This means that high frequency words will over time produce the lowest prediction error and are thus recognised faster and more accurately. Therefore, word frequency effects arise not as a consequence of manually changing resting activation levels but from the weights within the network changing according to the input distribution.

A final model of word recognition reviewed here is the Bayesian Reader Model \citep{norris2006bayesian}. This model not only accounts for the frequency effect but also aims to explain \textbf{why} frequency effects should arise in the first place. The model is a simple Bayesian model that integrates a word's prior probability \citep[for which][uses its frequency]{norris2006bayesian} with the incoming evidence. Thus, high frequency words are recognised faster than low frequency ones. According to \citet{norris2006bayesian, norris2013models} this constitutes an ``ideal observer'' model, solving the task at hand as optimally as possible. This explains not only why frequency effects should arise in the first place but also why they play out differently in different experiments.

To summarise, these five models offer three broad explanations of frequency effects. Serial search models explain them in terms of list ordering effects; the Logogen, interactive activation and triangle model propose network models where frequency is reflected in units' thresholds/activation levels or in connection weights; and finally, the Bayesian Reader proposes that word frequencies provide lexical priors that contribute to an optimal decision process in word recognition. 

Interestingly, reaction times for example in lexical decision are best predicted not by raw word frequencies but by log- or rank-transformed frequencies. Again, the various models account for this in different ways. Since the serial search model incorporates word lists, it directly predicts a rank transformation of word frequencies \citep[see also][]{murray2004serial}. In the interactive activation model, the resting activation levels of the word units are set according to the words' log frequency \citep[Chapter 7]{mcclelland1989explorations}. In the triangle model, training items are sampled such that they have a probability proportional to $\log(\text{frequency} + 2)$. Note that \citet{seidenberg1989distributed} do this for practical rather than principled reasons, as sampling training items proportionally to full token frequencies would require orders of magnitude more computation to achieve the same coverage \citep[this is a practice adopted also in other work, see for instance][]{li2007dynamic}. Finally, the Bayesian Reader utilises raw word frequencies.

Comparison of these models highlights a few key questions about how to model the word frequency effect: first, how and why does the frequency effect arise in the first place? Does it arise naturally as a consequence of the input data? And what mechanism does the model provide for how the frequency differences are acquired? Secondly, how does the model keep track of word frequencies? Are there ``counters'' for each individual word \citep[see also][]{baayen2010demythologizing}? And finally, how does the model account for the fact that reaction times are best predicted by log- or rank-transformed frequencies rather than raw frequency counts?

We now turn to a more recent model of word comprehension and production, the Discriminative Lexicon Model \citep[DLM;][]{Baayen2018a, baayen2019discriminative}. This model provides a perspective on the mental lexicon in which mappings between numeric representations for form and meaning play a central role. This model conceptualizes comprehension as involving mappings from high-dimensional modality-specific form vectors to high-dimensional representations of meaning.  The initial stage of speech production is modeled as involving a mapping in the opposite direction, starting with a high-dimensional semantic vector (known as embeddings in computational linguistics) and targeting a vector specifying which phone combinations drive articulation. The DLM has been successful in modelling a range of different morphological systems \citep[e.g.][]{chuang2020estonian, chuang2021vector, denistia2021morphology, nieder2023discriminative, heitmeier2021modeling} as well as behavioural data such as acoustic durations \citep{schmitz2021durational, stein2021morpho, chuang2021vector}, (primed) lexical decision reaction times \citep{Gahl2022thyme, heitmeier2023trial} and data from patients with aphasia \citep{heitmeier2020simulating}. 

The DLM's mappings between form and meaning are implemented by means of matrices. This general matrix-based approach is referred to as linear discriminative learning (LDL).  LDL can be implemented in two ways: by means of the matrix algebra underlying multivariate multiple regression (henceforth the `endstate learning', EL), or by means of incremental regression using the error-driven learning rule of \citet{Widrow1960} (henceforth WHL).  EL is computationally efficient, WHL is computationally demanding.  Conversely, WHL is sensitive to the frequencies with which words are presented for learning, whereas EL is fully type-based (i.e. words' token frequencies do not play a role).  

Thus, the DLM proposes that frequency effects arise due to the distribution of the input data: higher frequency words occur more often in the input data, and therefore, the prediction error will be smallest for high frequency words \citep[see][for studies utilising WHL to obtain frequency-informed mapping matrices]{chuang2021bilingual, heitmeier2021modeling}. Word frequencies are not stored explicitly; rather, they have effects on the weights in the mappings. This account is similar to how frequency effects arise in the triangle model. Similar to the triangle model, the DLM also suffers from computational issues: training on the full frequency distribution with WHL is computationally very demanding. 

Recent modelling efforts with the DLM have been limited by the disadvantages of EL and WHL: they either had to opt for EL, which resulted in models that were not informed about word frequencies \citep[e.g.][]{heitmeier2023trial}, or for WHL, which limited the amount of data the models could be trained on \citep[e.g.][]{heitmeier2021modeling, chuang2021bilingual}. The present paper aims to solve this problem by introducing a new method for computing the mapping matrices that takes frequency of use into account but is computationally efficient by making use of a numerically efficiently solvable solution: ``Frequency-informed learning'' (FIL). FIL can be used instead of the already established WHL and EL to compute mapping matrices in the DLM.

In the following we compare the three different methods of estimating mapping matrices in the DLM. We show how the model is able to account for frequency effects using WHL and the newly introduced FIL. We demonstrate that FIL is equivalent to training the model incrementally on full token frequencies and is superior to utilising log-transformed frequencies. We show how the DLM is able to model reaction times linearly without the need of log- or rank-transformations. Finally, we investigate what role the order in which words are learned plays in word recognition.

This study is structured as follows: we first lay out the basic concepts underlying linear discriminative learning in Section~\ref{sec:ldl}.  In Section~\ref{sec:fil}, we discuss EL, and subsequently, WHL. Against this background, we proceed with proposing a new method computing non-incremental, yet frequency-informed learning (FIL).  We then present three case studies, one using FIL to model visual lexical decision latencies from the Dutch Lexicon Project \citep{keuleers2010practice} (Section~\ref{sec:dlp}), one where we use FIL to model spoken word recognition in Mandarin (Section~\ref{sec:mandarin}) and a third where we compare WHL and FIL in modelling first word learning with data from CHILDES \citep{brown1973first, demuth2006word} (Section~\ref{sec:childes}).  A discussion section brings this study to a close. 

\section{Linear Discriminative Learning: Basic concepts and notation}\label{sec:ldl}

In the DLM, word forms are represented as binary vectors that code the presence and absence of overlapping n-grams in the word form (e.g., \texttt{\#a}, \texttt{aa}, \texttt{ap}, \texttt{p\#} for the Dutch word \textit{aap}, `\textit{monkey}'). Form vectors are stored as row vectors in a `cue matrix' $\mathbf{C}$.  For an overview of how form vectors can be constructed, see \citet{heitmeier2021modeling}, and for form vectors derived from audio signals, see \citet{shafaei2021ldl}. Semantics are represented as real-valued vectors, following distributional semantics \citep{Landauer1998}, which are  stored as row vectors in a semantic matrix $\mathbf{S}$. 

To model comprehension, a mapping matrix $\mathbf{F}$ transforms the form vectors in $\mathbf{C}$ into the semantic  vectors in $\mathbf{S}$. Conversely, a production matrix $\mathbf{G}$ maps meanings onto forms. The matrices $\mathbf{F}$ and $\mathbf{G}$ are estimated by solving
\begin{eqnarray*}
\mathbf{CF} &=& \mathbf{S} \\
\mathbf{SG} &=& \mathbf{C} 
\end{eqnarray*}
where $\mathbf{CF}$ and $\mathbf{SG}$ refer to the matrix multiplication of $\mathbf{C}$ and $\mathbf{F}$ and of $\mathbf{S}$ and $\mathbf{G}$ respectively. Further information on this operation can be found for instance in 
\citet{Beaumont:1965}.
Given $\mathbf{F}$ and $\mathbf{G}$, we can estimate the predicted semantic vectors
$$
\hat{\mathbf{S}} = \mathbf{CF}
$$ 
and the predicted form vectors 
$$
\hat{\mathbf{C}} = \mathbf{SG},
$$ 
with the hat on $\mathbf{S}$ and $\mathbf{C}$ indicating that the predicted matrices are estimates (in the statistical sense) that approximate the gold standard vectors but will usually not be identical to these vectors.  It is often convenient to focus on individual words, in which case we have that 
\begin{eqnarray*}
\hat{\mathbf{s}} &=& \mathbf{cF} \\
\hat{\mathbf{c}} &=& \mathbf{sG}.
\end{eqnarray*}
To evaluate the accuracy of a comprehension mapping, the predicted semantic row vectors $\hat{\mathbf{s}}$ of $\hat{\mathbf{S}}$ are correlated with all the corresponding semantic row vectors in the gold standard semantic matrix $\mathbf{S}$. If the predicted semantic vector of a word form is closest to its target vector, it is counted as correct. This accuracy measure is referred to as correlation accuracy in the following, and we will sometimes denote it as \texttt{accuracy@1}\footnote{As only 2.7\% of the words in our Dutch dataset are homographs, for simplicity we use `strict evaluation' throughout this study, i.e. we do not take homographs into account while evaluating accuracy \citep[further information on the various methods to evaluate accuracy in][]{heitmeier2021modeling}.}. More lenient accuracy measures \texttt{accuracy@k} accept model performance as satisfying when s is among the top k nearest target semantic vectors of the predicted semantic vector $\hat{\mathbf{s}}$. For detailed introductions to the DLM, see \citet{baayen2018inflectional, baayen2019discriminative, heitmeier2021modeling, heitmeier2023linear}.

\section{Three methods for computing mappings in the DLM}\label{sec:fil}

This section introduces the two existing methods for computing mappings in the DLM, Endstate Learning (EL) and Widrow-Hoff learning (WHL), and explains their respective disadvantages using a small Dutch dataset from \citet{ernestus2003predicting}. Finally, we present Frequency-informed learning (FIL). Since it is much more computationally efficient than WHL, we also demonstrate its usage on a larger dataset from \citet{keuleers2010practice}. For expositional simplicity, we focus mainly on comprehension.  We note, however, that frequency-informed mappings are equally important for modeling production.

\subsection{Setup}

\subsubsection{Data}\label{sec:data}

For the present section, we used two datasets:

\textbf{Small dataset:} A subset of 2,646 singular and plural Dutch nouns and verbs (for which frequency was at least 1) taken from a dataset originally extracted from the Dutch CELEX database \citep{baayen1995celex} by  \citet{ernestus2003predicting}\footnote{Data, code and statistical models presented in this study are available at \url{https://osf.io/h2szj/}}. These words have monomorphemic stems ending in an obstruent that is realized as voiceless when word-final but that in syllable onset appears with voicing in some words and without voicing in others. We used 300-dimensional Dutch fasttext embeddings \citep{grave2018learning} to represent semantics\footnote{\url{https://fasttext.cc/docs/en/crawl-vectors.html\#models}}. Since we were unable to obtain word embeddings for all words in our dataset, this left us with 2,638 word forms (i.e. excluding 8 word forms).

\textbf{Large dataset:} We also present results with a larger dataset extracted from the Dutch Lexicon Project \citep[DLP,][]{keuleers2010practice} in later sections. We used all 13,669 words from the DLP for which we were able to obtain fasttext embeddings. 

The frequencies of use that we use in this study when working both with the small and the large datasets are taken from CELEX. 

\subsubsection{Modelling choices}\label{sec:choices}

In what follows, we present results with two different form vector setups:

\textbf{Low-dimensional form vectors:} We  make use of bi-grams for representing word forms (used in their orthographic representation), resulting in a dimensionality of 360. The use of bigrams is motivated by the choice to minimize the carbon footprint and duration of our simulations, especially when using WHL.  

\textbf{High-dimensional form vectors:} For the new, computationally highly efficient frequency-sensitive model that is at the heart of this study, we made use of trigrams. For the small dataset this resulted in a form vector dimensionality of 1,776, and 4,678 for the large dataset.

\subsection{Endstate Learning}

\noindent
The first implementation of the DLM in the R \citep{rcoreteam2020r} package \textbf{WpmWithLdl} \citep{Baayen2018a} estimates the `endstate' of learning.  This  implementation constructs a mapping $\mathbf{F}$ between the cue  matrix $\mathbf{C}$ and the semantic matrix $\mathbf{S}$ using the pseudo-inverse by solving the following set of equations:

\begin{align}
    \mathbf{CF} &= \mathbf{S} \nonumber\\
    \mathbf{C}^T \mathbf{C F} &= \mathbf{C}^T \mathbf{S} \nonumber\\
    (\mathbf{C}^T \mathbf{C})^{-1} \mathbf{C}^T \mathbf{C F} &= (\mathbf{C}^T \mathbf{C})^{-1} \mathbf{C}^T \mathbf{S} \nonumber\\
    \mathbf{IF} &= (\mathbf{C}^T \mathbf{C})^{-1} \mathbf{C}^T \mathbf{S} \nonumber\\
    \mathbf{F} &= (\mathbf{C}^T \mathbf{C})^{-1} \mathbf{C}^T \mathbf{S} \label{eq:pseudo}
\end{align}
\noindent
where $\mathbf{I}$ denotes the identity matrix and $\mathbf{C}^T$ the transpose of $\mathbf{C}$. Details on these equations (known as the normal equations in statistics) can be found in \citet{Faraway:2005}.

Computing the pseudo-inverse as implemented in  \textbf{WpmWithLdl} is expensive and prohibitively so for larger datasets. Fortunately, there now exists a very efficient method implemented in the Julia \citep{bezanson2017julia} package \textbf{JudiLing}\footnote{\url{https://github.com/MegamindHenry/JudiLing.jl}} that makes use of the  Cholesky decomposition \citep{luo2021judiling, heitmeier2023linear}. This, together with additional speed-ups due to Julia being in general faster than R and the use of sparse matrices means that \textbf{JudiLing} can handle much larger datasets compared to \textbf{WpmWithLdl} \citep{luo2021judiling}.

The endstate learning (EL) results in optimal mapping matrices that reduce the error between the predicted and the target vectors as much as possible, for all word forms. It is optimal in the least-squares sense, and the underlying mathematics are identical to that of multivariate multiple regression.  This method is characterized as estimating the `endstate' of learning, because the mappings it estimates can also be approximated by using incremental learning (WHL) applied to an infinite number of passes through one's dataset (assuming that each word occurs a single time in the dataset). With infinite experience, every word has been experienced an equal and infinite amount of times.  Any effects of frequency of occurrence in this model are an epiphenomenon of lexical properties such as word length and neighborhood density \citep{Nusbaum:85,Baayen:2001}.

\begin{figure}[!htb]
    \centering
    \includegraphics[width=.8\textwidth]{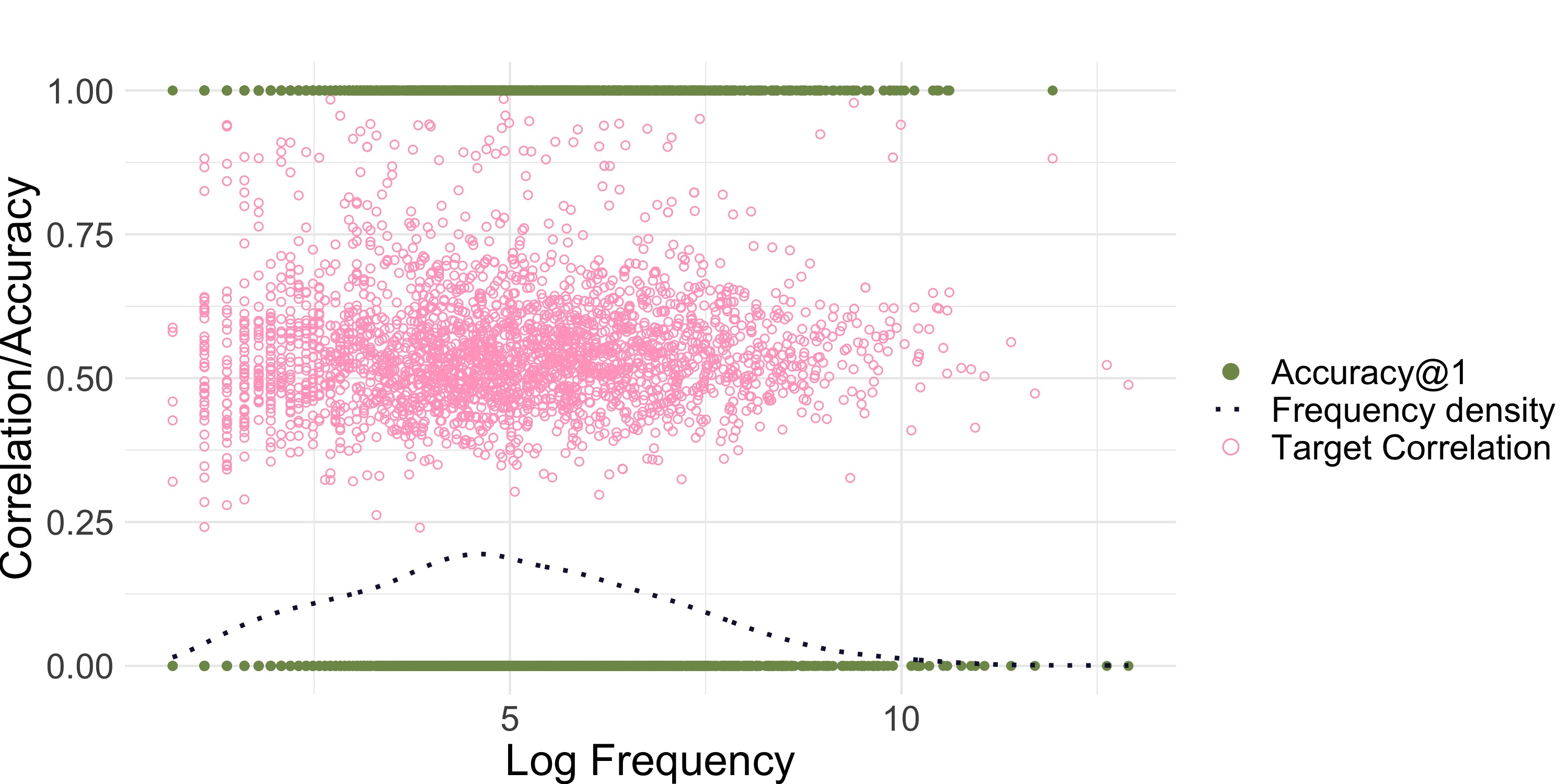}
    \caption{Endstate learning. The green filled dots on the horizontal lines at 0 and 1 represent the correlation accuracies@1 for the individual words (counted as correct if the semantic vector most correlated with the predicted semantic vector is the target), and the light pink circles represent the correlation values of words' predicted semantic vectors with their target vectors.  The dark blue dotted line presents the estimated kernel density for log frequency. \\
    There is no discernible relationship between Log Frequency and correlation/accuracy for endstate learning.}
    \label{fig:eol}
\end{figure}

Figure~\ref{fig:eol} illustrates the frequency-free property of EL.  It presents the results obtained with an EL model with low-dimensional form vectors for the small dataset introduced in the preceding section, modeling comprehension with the matrix $\mathbf{F}$.  The average correlation accuracy for this model is low, at 40.8\%;  below, we will show how this accuracy can be improved to 83\% by increasing the dimensionality of the form vectors.

To illustrate the absence of a relationship between word frequency and correlation accuracy, we used a binomial generalised linear model\footnote{\url{https://github.com/JuliaStats/GLM.jl}} with a logit link function, modeling the probability of correct recognition as a function of log frequency (here and in all later analyses, before log-transformation, a backoff value of 1 was added to word frequencies).  Log frequency was not a good predictor of accuracy ($p = 0.9772$).

In summary, one important advantage of endstate learning is that it is can be computed very quickly: on a MacBook Pro (2017) with a 3.1 GHz Quad-Core Intel Core i7 processor and for the small dataset with low-dimensional form vectors it takes about 20 milliseconds.
A second important advantage of this method is that it is frequency-free: it shows what can be learned when frequency of use is not allowed to play a role.  In other words, the EL probes the quantitative structure of a dataset, under the assumption that usage can be ignored.  Thus, the EL method dovetails well with generative approaches to morphology that deny any role to usage in grammar and work with types only.

However, this is also the achilles heel of the EL method: EL is purely type-based and is blind to the consequences of token frequency for learning. Well-established effects of frequency of use \citep[see, e.g.,][]{Baayen:Dijkstra:Schreuder:1997,bybee2001frequency} are not captured.  This is why an alternative way of estimating mappings between form and meaning is implemented in the \textbf{JudiLing} package: incremental learning.

\subsection{Incremental learning}

\noindent
Instead of making use of the efficient method estimating the endstate of learning, we can also learn the mappings incrementally using the Widrow-Hoff learning rule \citep{Widrow1960}, a form of error-driven learning closely related to the Rescorla-Wagner learning rule (\citeauthor{Rescorla1972}, \citeyear{Rescorla1972}; for a discussion of the Widrow-Hoff learning rule in language research see also \citeauthor{milin2020keeping}, \citeyear{milin2020keeping}). Here, the idea is that each time a word is encountered, a learning event occurs, and the mappings between form and meaning are updated in such a way that next time the same word is encountered (if no unlearning occurs intermittently) the mappings will be more accurate.  To be precise, instead of obtaining the mapping matrix $\mathbf{F}$ via equation \ref{eq:pseudo}, it is learned gradually via the following equation:
\begin{align*}
    \mathbf{F}_{t+1} = \mathbf{F}_t + \mathbf{c}_t^T \cdot (\mathbf{s}_t - \hat{\mathbf{s}}_t) \cdot \eta
\end{align*}
where $\mathbf{F}_t$ is the state of the matrix $\mathbf{F}$ at learning step $t$, $\mathbf{c}_t$ and $\mathbf{s}_t$ are the form and semantic vectors of the wordform encountered at $t$, and $\hat{\mathbf{s}}_t = \mathbf{c}_t \cdot \mathbf{F}_t$. How fast learning takes place is controlled via the learning rate parameter $\eta$. High learning rates lead to fast learning and unlearning, whereas low learning rates result in slower learning and less unlearning.

\begin{figure}[!htb]
     \centering
    \includegraphics[width=\textwidth]{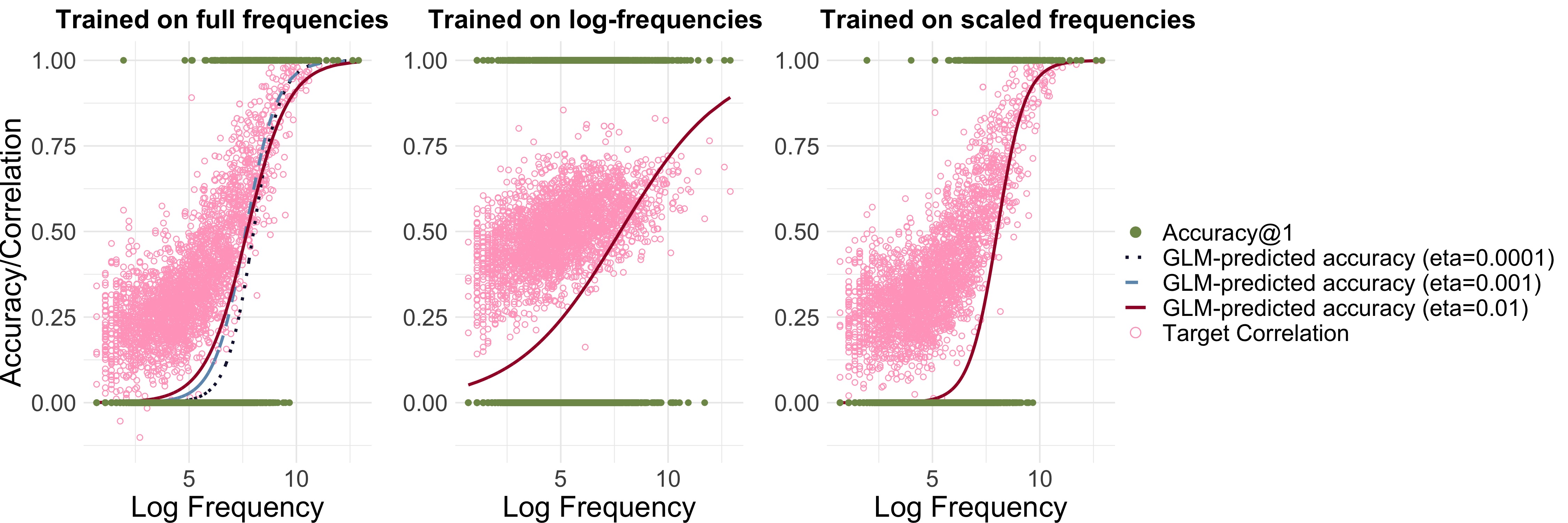}
        \caption{Relationship between accuracy and frequency for incremental learning. \textbf{Left panel:} Mapping trained using full frequencies. Predicted accuracy is depicted for three different learning rates ($\eta \in \{0.01, 0.001, 0.0001\}$), and the light pink circles present target correlations for $\eta=0.01$. \textbf{Center panel:} Mapping trained using log-transformed frequencies. \textbf{Right panel:} Mapping trained using frequencies divided by a factor of 100. \\
        While there is a strong relationship between log frequency and accuracy/correlation when training on full frequencies and scaled frequencies, this relationship is attenuated when training on log-transformed frequencies.}\label{fig:incr}
\end{figure}

Incremental learning has the advantage that we can model learning in a frequency-informed way by translating frequencies into learning events. For example, if a word has a frequency of 100, it is presented for learning 100 times. We used the Widrow-Hoff learning rule as implemented in \textbf{JudiLing} to incrementally learn the mapping matrix $\mathbf{F}$. This mapping is now frequency-informed. We experimented with low-dimensional form vectors and three different learning rates: 0.01, 0.001, and 0.0001, modelling words in the small dataset. The average correlation accuracy for the three simulations was 15.5\%, 14.0\% and 10.1\% respectively. As can be seen in the left panel of Figure~\ref{fig:incr}, when learning the mappings incrementally according to their frequency, a clear relationship between frequency and accuracy is obtained for all learning rates: the more frequent a word is, the more accurately it is learned. A binomial GLM indicated a highly significant relationship between frequency and accuracy ($p<0.001$ for all learning rates). The lower the learning rate, the steeper the increase in accuracy: accuracy is lower for low-frequency words and higher for high-frequency ones.

There are two disadvantages to this approach, one practical, and the other theoretical. The theoretical problem is that for many datasets, there is no intrinsic order in which words are learned. For the present dataset, which is basically a word frequency list, we do not have available any information about the order in which Dutch speakers encounter these words over their lifetime. It is only for intrinsically ordered data, such as child directed speech in corpora such as CHILDES \citep{macwhinney2014childes}, that incremental learning comes into its own (see Section~\ref{sec:childes}).  The practical problem is that updates with the Widrow-Hoff learning rule are computationally expensive. For the present dataset, estimating the mapping matrix $\mathbf{F}$ took $\approx$25 minutes on a MacBook Pro (2017) with a 3.1 GHz Quad-Core Intel Core i7 processor, even though the use of bi-grams resulted in a form dimensionality that was already too small to obtain good accuracy. If we would use the better-suited tri-grams (i.e. the high-dimensional form vectors described in Section~\ref{sec:choices}), the estimated time for computing the mapping matrix $\mathbf{F}$ increases strongly even in an optimized language such as Julia.

The computational cost of WHL may be alleviated in (at least) two ways. One option is to transform frequencies by taking logarithms \citep[see, e.g.,][]{seidenberg1989distributed}. The resulting average correlation accuracy is 26.8\%. The relationship between frequency and accuracy can be seen in the center panel of Figure~\ref{fig:incr}. While the log transformation does indeed reduce computational costs, it is questionable whether such a transformation is justified and realistic. Low frequency words become proportionally more accurate, while high frequency ones become less so. If WHL with empirical frequencies for time-ordered learning events is taken as a gold standard, then a log-transformation distorts our estimates considerably.

A second option is to simply scale down frequencies by dividing them by a fixed number. By applying a ceiling function to the result, we avoid introducing zero frequencies. Training the model using frequencies divided by 100 speeds up the learning to $\approx$12s and does not distort the learning curve (see right panel of Figure~\ref{fig:incr}). The disadvantage of this method is that there are far fewer learning events. As a consequence, words are learned less well. Accordingly, the average correlation accuracy drops to 10.3\%.

In summary, it is in principle possible to estimate mapping matrices with incremental learning. This is theoretically highly attractive for data that are intrinsically ordered in learning time \citep[see, e.g.,][for the modeling of within-experiment learning]{heitmeier2023trial}.  For unordered data, some random order can be chosen, but for larger datasets, it would be preferable to have a method that is agnostic about order but nevertheless accounts in a principled way for the consequences of experience for discriminative learning.

\subsection{Non-incremental, yet frequency-informed mappings}

A solution to this conundrum is to construct frequency-informed mappings between form and meaning. Thinking back to incremental learning, learning a word $w_i$ with frequency count $f_i$ involved learning the mapping from a cue vector $\mathbf{c}_i$ to the word's meaning $\mathbf{s}_i$ $f_i$ times. We could thus construct matrices $\mathbf{C}_f$ and $\mathbf{S}_f$ reflecting the entire learning history: $\mathbf{C}_f$ and $\mathbf{S}_f$ are $\mathbf{C}$ and $\mathbf{S}$ with word forms $w_i$ and semantic vectors $\mathbf{s}_i$ repeated according to their frequency count $f_i$. We are looking for the mapping $\mathbf{F}_f$ and $\mathbf{G}_f$ such that 
$$\mathbf{S}_f = \mathbf{C}_f \mathbf{F}_f$$
$$\mathbf{C}_f = \mathbf{S}_f \mathbf{G}_f$$

Formally, let $\mathbf{C} = \begin{pmatrix} \mathbf{c}_1 & \mathbf{c}_2 & ... & \mathbf{c}_m\end{pmatrix}^T \in \mathbb{R}^{m \times r}$  and $\mathbf{S} = \begin{pmatrix} \mathbf{s}_1 & \mathbf{s}_2 & ... & \mathbf{s}_m\end{pmatrix}^T \in \mathbb{R}^{m \times q}$ where each word $w_i$ of the $m$ wordforms corresponds to a row in the two matrices with cue vector $\mathbf{c}_i$ and semantic vector $\mathbf{s}_i$. Each word form $w_i$ has a frequency count $f_i$. 

We can create two new matrices $\mathbf{C}_f$ and $\mathbf{S}_f$ where the cue and semantic vectors of the wordforms are repeated according to their frequency count $f_i$. We want to find the mapping matrix $\mathbf{F}_f$ mapping from $\mathbf{C}_f$ to $\mathbf{S}_f$. We use the following solution for computing the mapping matrix \citep[see also][]{baayen2018inflectional}:
\begin{align}
    \mathbf{F}_f = &(\mathbf{C}_f^T \mathbf{C}_f)^{-1}\mathbf{C}_f^T \mathbf{S}_f\label{eq:line1}\\
    \overset{\text{see Supplementary}}{=} &\left(\sum_{i=1}^m f_i\mathbf{c}_i\mathbf{c}_i^T\right)^{-1}\left(\sum_{i=1}^m f_i\mathbf{c}_i\mathbf{s}_i^T\right)\label{eq:line2}\\
    = &\left(\sum_{i=1}^m \sqrt{f_i}\mathbf{c}_i\left(\sqrt{f_i}\mathbf{c}_i\right)^T\right)^{-1}\left(\sum_{i=1}^m \sqrt{f_i}\mathbf{c}_i\left(\sqrt{f_i}\mathbf{s}_i\right)^T\right)\label{eq:line3}\\
    \overset{\text{see Supplementary}}{=} &\left(\sum_{i=1}^m \sqrt{\frac{f_i}{k}}\mathbf{c}_i\left(\sqrt{\frac{f_i}{k}}\mathbf{c}_i\right)^T\right)^{-1}\left(\sum_{i=1}^m \sqrt{\frac{f_i}{k}}\mathbf{c}_i\left(\sqrt{\frac{f_i}{k}}\mathbf{s}_i\right)^T\right)\label{eq:line4}
\end{align}
with a constant $k \in \mathbb{R}_{>0}$. Since $k$ does not change the solution, we can set it such that the algorithm is numerically more stable, for example to $k=\max_{j\in1:m}f_j$ so that we have $p_i = \frac{f_i}{\max_{j\in1:m}f_j}$ and therefore
\begin{align}
    \mathbf{F}_f =\left(\sum_{i=1}^m \sqrt{p_i}\mathbf{c}_i\left(\sqrt{p_i}\mathbf{c}_i\right)^T\right)^{-1}\left(\sum_{i=1}^m \sqrt{p_i}\mathbf{c}_i\left(\sqrt{p_i}\mathbf{s}_i\right)^T\right).
\end{align}
Now let $\mathbf{P} \in \mathbb{R}^{m \times m}$ be a diagonal matrix with $p_{ii} = p_i$ for $i \in 1,...,m$. Then we can define $\Tilde{\mathbf{C}} = \sqrt{\mathbf{P}}\mathbf{C}$  and $\Tilde{\mathbf{S}} = \sqrt{\mathbf{P}}\mathbf{S}$ so that $\Tilde{\mathbf{c}}_i = \sqrt{p_i}\mathbf{c}_i$ and $\Tilde{\mathbf{s}}_i = \sqrt{p_i}\mathbf{s}_i$. Then we have
\begin{align*}
    \mathbf{F}_f = &\left(\sum_{i=1}^m \Tilde{\mathbf{c}}_i\Tilde{\mathbf{c}}_i^T\right)^{-1}\left(\sum_{i=1}^m \Tilde{\mathbf{c}}_i\Tilde{\mathbf{s}}_i^T\right)\\
    = &(\Tilde{\mathbf{C}}^T \Tilde{\mathbf{C}})^{-1}\Tilde{\mathbf{C}}^T \Tilde{\mathbf{S}}
\end{align*}
Therefore, the pair $(\Tilde{\mathbf{C}}, \Tilde{\mathbf{S}})$ has the same mapping matrices as $(\mathbf{C}_f, \mathbf{S}_f)$.

Practically, this means that we can first weigh $\mathbf{C}$ and $\mathbf{S}$ with the pertinent frequencies to obtain $\Tilde{\mathbf{C}}$ and $\Tilde{\mathbf{S}}$. We can then use the solution in equation \ref{eq:pseudo} (making use of algorithms such as Cholesky decomposition) to obtain frequency-informed mappings between these two matrices.\footnote{Note that while we here work with a learning rule for continuous vectors, FIL is also applicable to mappings between discrete vectors, as in Rescorla-Wagner learning \citep{rescorla1967pavlovian} which is used in Naive Discriminative Learning \citep[NDL,][]{baayen2011amorphous}:  instead of having to impose some random order on the learning events, the expected value irrespective of order can be estimated using FIL.} 

In what follows, we sketch the new possibilities enabled by this method, to which we will refer as frequency-informed learning (FIL). A first, practical, advantage of FIL is that it is efficient and fast. A second, theoretical, advantage is that predictions are available for datasets for which no information about the order of learning is available.

\subsubsection{Low-dimensional modeling}

We first consider modeling studies using the low-dimensional vectors that we used in the preceding sections together with the small dataset. We chose this low dimensionality in order to avoid long computation times for WHL for these exploratory studies. 

\begin{figure}[!htb]
     \centering
         \includegraphics[width=.8\textwidth]{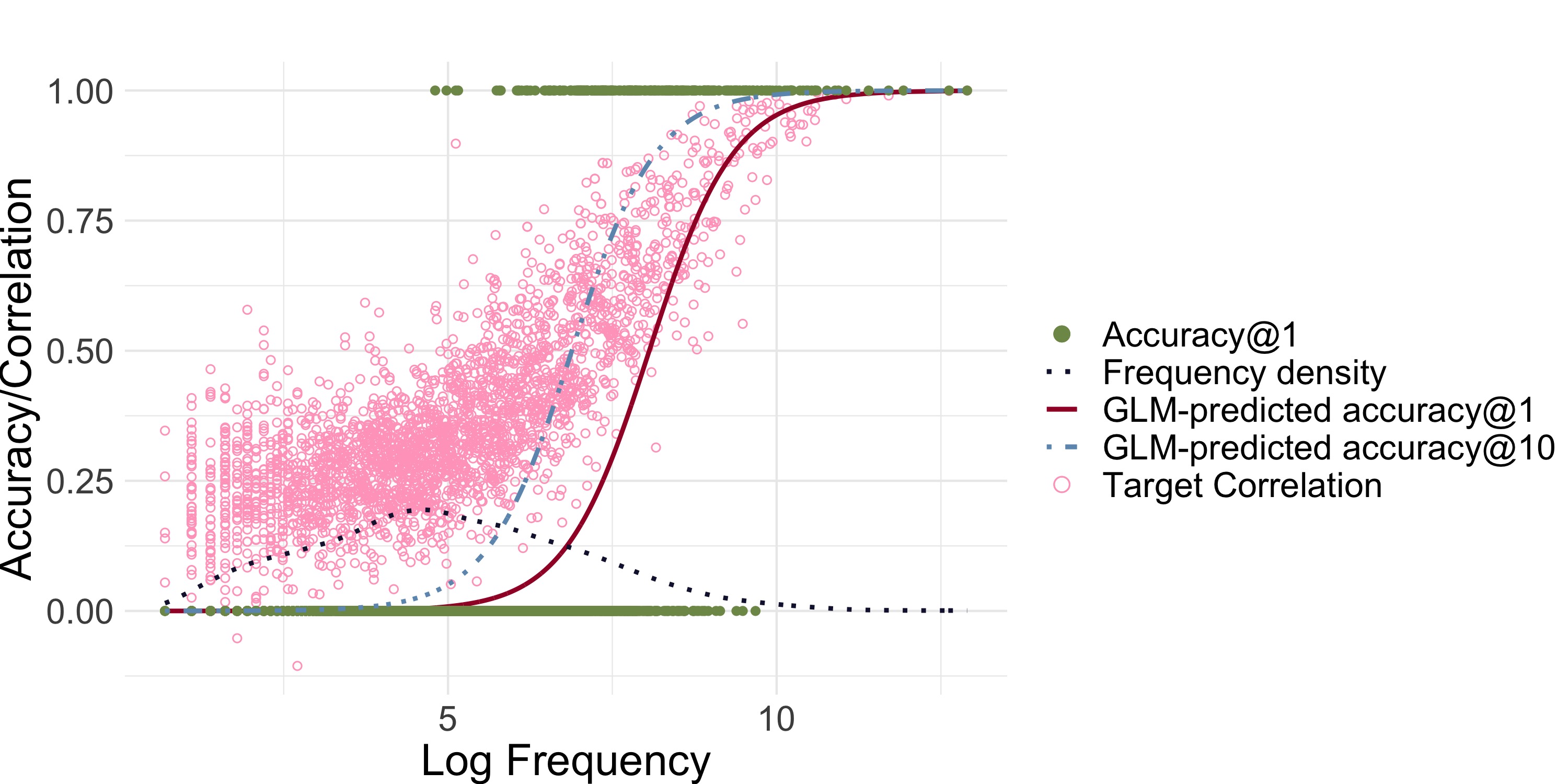}
         \caption{Frequency-informed learning. The red solid line presents the predictions of a GLM when a success is defined as the predicted vector being the closest to its gold standard target vector in terms of correlation (\texttt{accuracy@1}).  The light blue dashed line represents model predictions when a success is defined as the correlation being among the top 10 (\texttt{accuracy@10}).  The dark blue dotted line visualizes the estimated density of the log-transformed frequencies.  The green filled dots represent the successes and failures for \texttt{accuracy@1}.  The light pink circles represent for each word the correlation of the predicted and gold-standard semantic vectors.\\
         There is a strong relationship between log frequency and correlation/accuracy, and the GLM-predicted \texttt{accuracy@10} is shifted to the left, i.e. \texttt{accuracy@10} rises for lower frequencies.}
         \label{fig:fi_norm}
\end{figure}

Figure~\ref{fig:fi_norm} shows the relationship between log-transformed frequency and accuracy predicted by a logistic GLM regressing the correctness of FIL responses on log frequency. \texttt{Accuracy@k} is set to 1 if a word's target semantic vector is among the $k$ target vectors that are most correlated with the predicted semantic vector, and to 0 otherwise.

Correlation and \texttt{accuracy@1} increase for higher frequency, as required.  When comparing accuracies with the frequency distribution depicted in the same plot, we can also see that there is a large number of low frequency words with very low (predicted) accuracy. Although for most words, the correlations are relatively high, the overall \texttt{accuracy@1} is low, at 9.9\%.  When we relax our criterion for accuracy, using \texttt{accuracy@10},  counting a predicted vector as correct if the target vector is among the 10 closest semantic vectors, we see that the accuracy starts to rise earlier, but there is still a significant portion of words for which even \texttt{accuracy@10} is zero.

\begin{figure}[!htb]
    \centering
    \includegraphics[width=.8\textwidth]{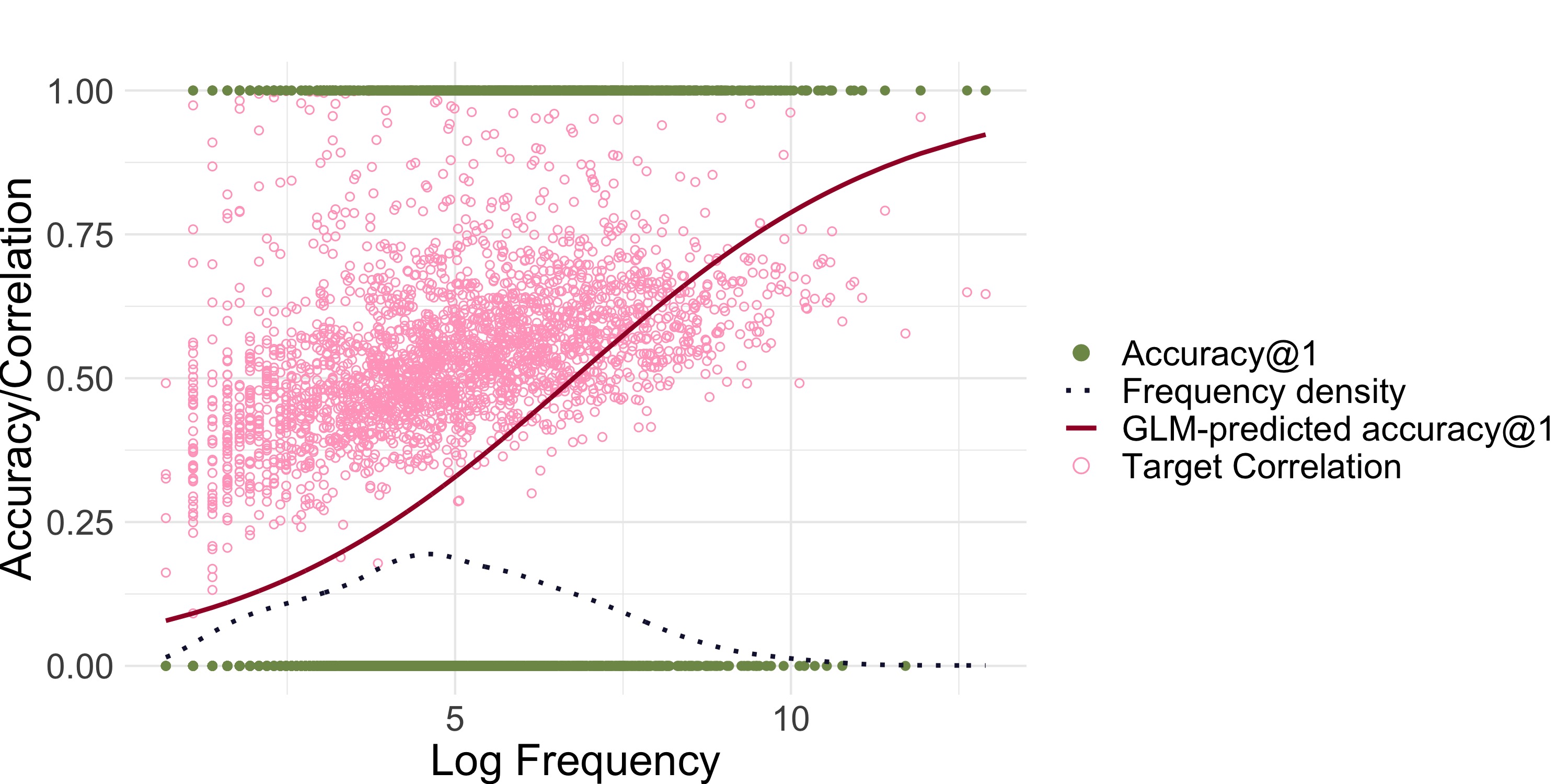}
    \caption{\texttt{Accuracy@1} as a function of log frequency, using frequency-informed learning with log-transformed frequencies. \\
    When FIL is trained with log-transformed frequencies, lower-frequency words are recognized more accurately, but higher-frequency words less accurately.}
    \label{fig:fi_log}
\end{figure}

The relatively large number of words with very low accuracy raises the question of whether accuracy can be improved by using log-transformed frequencies for FIL.  Figure~\ref{fig:fi_log} clarifies that accuracies increase for lower-frequency words, but decrease somewhat for higher-frequency words.  The average \texttt{accuracy@1} is accordingly higher at 34.6\%.

The upper half of Table~\ref{tab:comparison} provides an overview of the \texttt{accuracies@1} for different combinations of learning (incremental/frequency-informed) and kind of frequency used (untransformed, scaled, or log-transformed).  Here, we observe first of all that  endstate learning offers the highest accuracy (40.8\%), followed by log-frequency informed learning (34.6\%).  

\begin{table}[!htb]
\small
    \centering
    \begin{tabular}{|lrr|} \hline
            \multicolumn{3}{|c|}{\textbf{low-dimensional form vectors}} \\ \hline
        Model & Average \texttt{accuracy@1} & Frequency-weighted \texttt{accuracy@1} \\
        \hline
    Endstate learning (EL) & 40.8\% & 32.2\%\\
        Incremental learning ($\eta =0.01$) & 15.5\% & 78.1\%\\ 
        Incremental learning ($\eta =0.001$) & 14.0\% & 79.1\%\\ 
        Incremental learning ($\eta =0.0001$) & 10.1\% & 75.6\%\\ 
        Incremental learning log-frequencies & 26.8\%& 65.4\%\\
        Incremental learning scaled frequencies & 10.3\%& 75.3\%\\ 
        Frequency-informed learning (FIL) & 9.9\%& 74.9\%\\
        Log-frequency-informed learning & 34.6\% & 71.3\%\\
        \hline
        \multicolumn{3}{|c|}{\textbf{high-dimensional form vectors}} \\ \hline
        Model/Dataset size & Average \texttt{accuracy@1} & Frequency-weighted \texttt{accuracy@1} \\
        \hline
        
        Endstate learning (EL)/Small & 83.0\% & 79.8\%\\
        Frequency-informed learning (FIL)/Small & 22.3\% & 89.4\%\\
         \hline
         Endstate learning (EL)/Large & 67.8\% & 43.7\%\\
        Frequency-informed learning (FIL)/Large & 5.1\% & 79.8\%\\
         \hline
    \end{tabular}
    \caption{Comparison of average and frequency-weighted \texttt{accuracy@1} (the term ``frequency-weighted accuracy'' is introduced in Section~\ref{sec:frequency_weighted}) across simulation studies. The small dataset contained 2,638 word forms, the large dataset 13,669. When working with low-dimensional form vectors, the small dataset was used throughout.}
    \label{tab:comparison}
\end{table}

\begin{figure}[!htb]
     \centering
    \includegraphics[width=\linewidth]{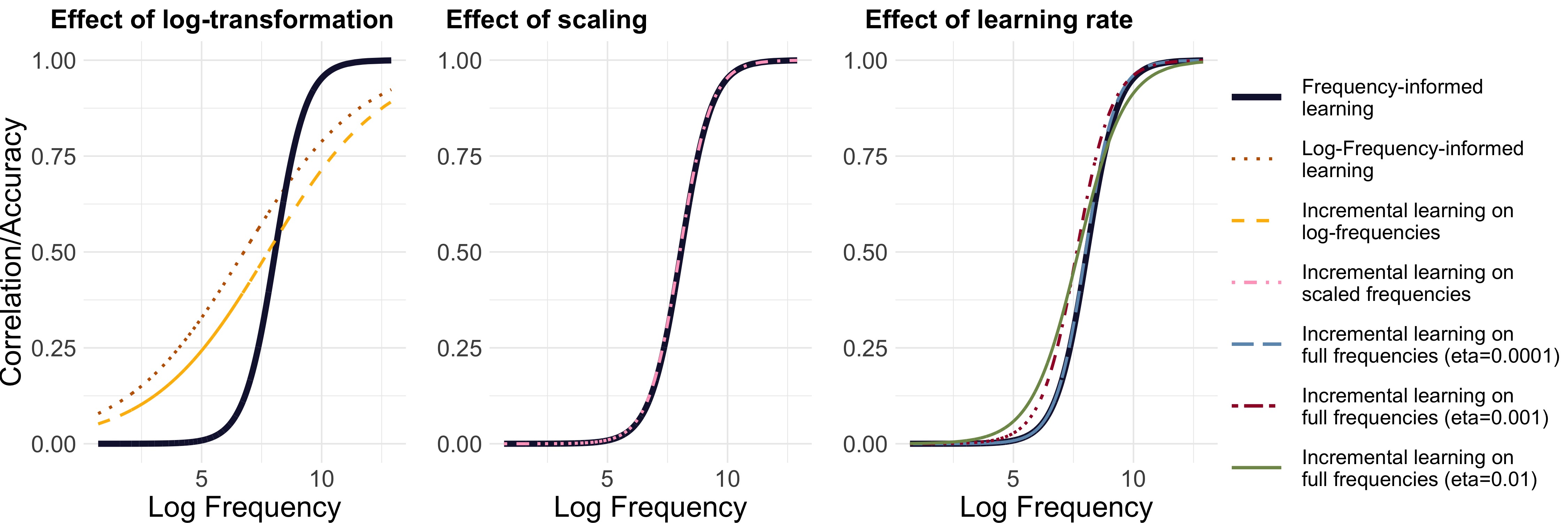}
        \caption{Comparison of methods.  GLM-predicted \texttt{Accuracy@1} with frequency-informed learning is plotted as a black line: The left panel compares methods based on log-frequencies, the center panel compares methods based on scaled frequencies and the right panel compares incremental learning with different learning rates. Incremental learning with scaled frequencies or with a very low learning rate ($\eta=0.0001$) is closest to frequency-informed  learning.}\label{fig:comparison}
\end{figure}

Figure~\ref{fig:comparison} highlights the differences between the model set-ups, comparing with FIL the effect of a log-transformation (left panel), of scaling (center panel), and of the learning rate (right panel).  It is noteworthy that frequency-informed learning with log-transformed frequencies departs the most from both FIL and incremental learning, which suggests that training on log frequency may artefactually increase learning performance.

Secondly, it can be observed that the incremental learning based on scaled frequencies is closest to frequency-informed learning in terms of average \texttt{accuracy@1}, as well as to incremental learning with the lowest learning rate. This suggests a) that scaling frequencies has a similar effect as lowering the learning rate in incremental learning and b) that frequency-informed learning approximates incremental learning for very low learning rates.

\subsubsection{High-dimensional modeling}

Importantly, the mappings that we used thus far are suboptimal: the dimensionality of the semantic vectors was small and the use of bi-grams for the form vectors often shows underwhelming performance \citep[see][for further discussion of the underlying reasons]{heitmeier2021modeling}.  While opting for low dimensionality decreased the computational costs for incremental learning immensely and was therefore necessary for comparing methods, we now proceed to investigate the accuracy of frequency-informed learning for larger, more discriminative cue matrices. To this end, we next experimented with the high-dimensional form vectors, still making use of the small dataset.

The model for the endstate of learning now performs much better, at an \texttt{accuracy@1} of 83\% instead of 40.8\%.  In other words, with infinite experience of just the words in this dataset, and with all token frequencies going to infinity, this is the best our simple multivariate multiple regression approach can achieve (conditional on the way in which we encoded form and meaning). 

\begin{figure}[!htb]
\centering
\includegraphics[width=.8\textwidth]{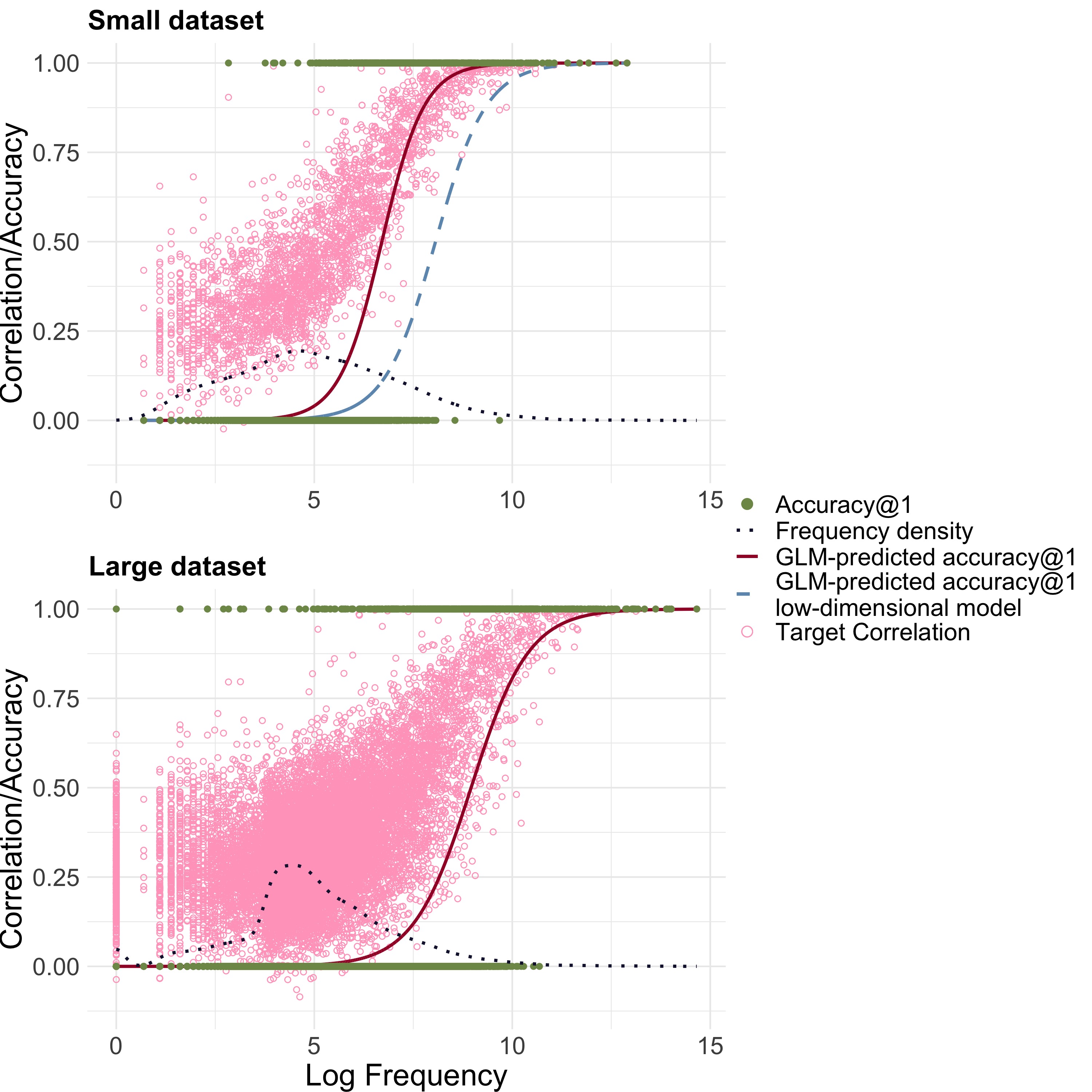}
    \caption{Predicted \texttt{accuracy@1} as a function of log frequency for high-dimensional representations of form and meaning (red solid line). The light blue dashed line shows the predicted accuracy based on the low-dimensional model, for comparison. The light pink circles represent the target correlations in the high-dimensional model. The small dataset refers to the dataset with 2,638 word forms based on \citet{ernestus2003predicting}, the large dataset to the dataset created from the DLP \citep{brysbaert2009moving} including 13,669 word forms. \\
    For the small and large datasets, a clear relationship between log frequency and correlation/accuracy is visible.}
    \label{fig:large_model}
\end{figure}

A model using FIL obtained an average \texttt{accuracy@1} of 22.3\%, which is clearly superior to the 9.9\% obtained for the lower-dimensional model.  The upper panel of Figure~\ref{fig:large_model} presents the predicted accuracy curves for the high-dimensional FIL model in red (solid line), and the low-dimensional FIL model in blue (dashed line). We see a rise in accuracy for lower frequencies.

\subsubsection{Increasing the dataset size}\label{sec:larger_dataset}
Having established the relationship between EL, WHL and FIL, we can also investigate how EL and FIL fare when the modelled dataset is significantly larger. To this end we used the large dataset introduced above. We found that accuracies in general were clearly lower: For EL, the \texttt{accuracy@1} was 67.8\% and for FIL it was 5.1\%. Qualitatively, the differences between EL and FIL are therefore similar (see also Figure~\ref{fig:large_model}), but our simple linear mappings clearly perform less well with very large datasets, especially when taking into account frequency.

\subsubsection{Frequency-weighted accuracy}\label{sec:frequency_weighted}
A final question is whether the way we have been evaluating accuracy so far is reasonable. In the average \texttt{accuracy@1} each word's accuracy contributes the same, that is, we have effectively calculated accuracy across all word types in our corpus. However, from a usage-based perspective, comprehending high frequency words is much more important than comprehending low frequency words --- this is why, for instance, second language learners are generally taught the most frequent words of a language first. In a usage-based analysis of our model we should therefore be calculating accuracy across word tokens instead. Practically this means that if we go through a corpus of written English, instead of counting how many unique words our model is able to comprehend, we count how many of all of the encountered word tokens are understood correctly.\footnote{As pointed out by a reviewer, this accuracy metric puts a lot of emphasis on high frequency closed class words which are often considered stopwords in computational linguistics (examples in English are \textit{the, of, and, at, by, of},...; see for instance the English stopword list in NLTK \citep{bird2006nltk}). However, stopwords have to be learned just as other words, so excluding them a-priori seems unprincipled. We also do not see a plausible cognitive mechanism that would filter out stopwords but not other high frequency open class words. Another issue is that what are considered stopwords in English or Dutch, are often incorporated inside word forms in other languages (e.g. what is expressed with prepositions in English is expressed with case inflection in Finnish). Excluding stopwords for English but including case inflections in Finnish would make modeling incomparable and incompatible across languages.}

Following this line of argumentation we also provide frequency-weighted accuracies in Table~\ref{tab:comparison}.  For instance, the word with the highest frequency in the Dutch Lexicon Project (large dataset) is \textit{de} (eng. \textit{the}). Since this word accounts for 7\% of all word tokens in the Dutch Lexicon Project (calculated by summing up the frequencies of all word types), it also contributes 7\% to our frequency-weighted accuracy measure. There are also 349 words with a frequency of 0 in CELEX (2.6\% of all word types) which accordingly do not contribute to the frequency-weighted accuracy at all. We find that with this method, the results flip in comparison with average \texttt{accuracy@1}: Generally, FIL and WHL based on untransformed frequencies perform the best, followed by methods based on log-transformed frequencies, while EL clearly performs the worst. For example, for the high-dimensional simulations on the large dataset, EL has a frequency-weighted \texttt{accuracy@1} of 43.7\%, while FIL achieves 79.8\%. This means that EL understands less than half of the word tokens it encounters in a corpus, while FIL comprehends about 8 out of 10 word tokens.

Whether type or token accuracy should be used to evaluate a model depends on the type of analysis conducted by the modeller. For a usage-based perspective as we haven taken here, a token-based measure is more appropriate. For other types of analysis, a type-based measure may be more suited. For instance, in the case of morphologically more complex languages such as Estonian or Russian, the modeller may well be interested in how well the model is able to understand and produce even low-frequency inflected forms of well-known lemmas. A token-based accuracy measure is not helpful in such cases.

To summarise, FIL provides an efficient way of estimating mappings between frequency-informed representations of form and meaning. FIL does not reach the average accuracy across types of EL, but, importantly, unlike for EL, accuracy varies systematically with frequency in a way similar to how human accuracy is expected to vary with frequency. Moreover, FIL clearly outperforms EL when accuracy is calculated across tokens rather than types. The question addressed in the next section is whether FIL indeed provides predictions that match well with a particular behavioral measure: reaction times in a visual and an auditory lexical decision task. 

\section{FIL-based modeling of reaction times}

\subsection{Visual word recognition in Dutch}\label{sec:dlp}

\noindent
In order to assess the possibilities offered by FIL-based modeling (using untransformed frequencies and the high-dimensional form and meaning vectors) for predicting behavioral measures of lexical processing, we return to the large dataset of reaction times to Dutch words described in Section~\ref{sec:data} (13,669 words represented as trigrams with a form dimensionality of 4,678, and semantics represented using 300-dimensional fasttext vectors).

\begin{figure}[!htb]
\centering
\includegraphics[width=0.6\textwidth]{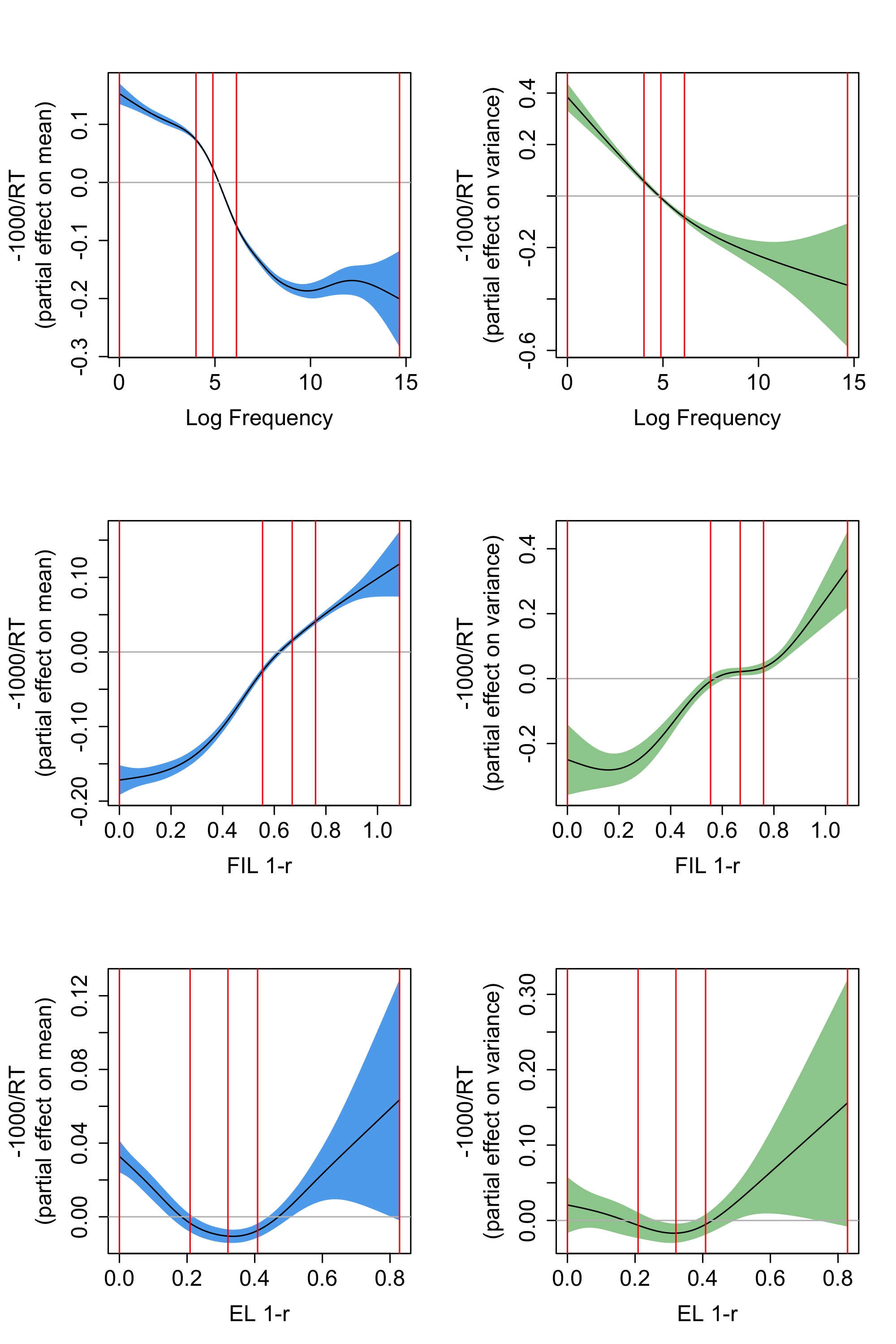}
\caption{
Partial effects for mean (left, confidence intervals in blue) and variance (right, confidence intervals in green, the y-axis is on the $\log(\sigma - 0.01)$ scale), for Gaussian Location-Scale GAMs predicting reaction times from log frequency (upper panels), from $1-r$ based on FIL (center panels), and from $1-r$ based on EL (bottom panels). The vertical red lines represent the 0\%, 25\%, 50\%, 75\%, and 100\% percentiles.\\
FIL 1-r is a solid predictor for mean and variance in RTs.
}
\label{fig:FIL_EL_RT_plot}
\end{figure}

Figure~\ref{fig:FIL_EL_RT_plot} presents the partial effects according to three Gaussian Location-Scale GAMs fitted to the response latencies in the Dutch Lexicon Project.  Response latencies were inverse transformed (-1000/RT) in order to avoid marked non-normality in the residuals.  (Effects for the untransformed RTs are very similar in shape, but confidence intervals are not reliable due to the marked departure from normality of the residuals). The left-hand panels present the partial effects for the mean (in blue), the right panels the partial effects for the variance (in green, on the $\log(\sigma-0.01)$ scale; for further information on how partial effects are calculated in GAMs see \citet{Wood:2017}).  The upper panels pertain to a GAM predicting RT from log frequency (AIC -16,384.8; a backoff value of 1 was again added before log-transformation of word frequencies). Mean and variance decrease non-linearly with increasing frequency.  The smooth for the mean shows the kind of non-linearity that is typically observed for reaction time data \citep[see, e.g.,][]{Baayen:2005,miwa2021nonlinearities}: the effect of log frequency levels off strongly for high-frequency words and to a lesser extent also for low-frequency words. The partial effect of the variance is less wiggly and decreases as frequency increases.  This decrease in variability for increasing frequency has at least two possible sources. First, high-frequency words are known to all speakers, whereas low-frequency words tend to be specialized and known to smaller subsets of speakers.  Second, more practice, as in the case of high-frequency words, typically affords reduced variability \citep[see, e.g.,][]{Tomaschek:Tucker:Baayen:2018}. 

A model-based measure that we expected to correlate with reaction time is the proximity of a word's predicted semantic vector to its corresponding gold standard vector (i.e. the target correlation). The more similar the two vectors are, the better a word's meaning is reconstructed from its form. In other words, the more effective a word form is in approximating its meaning, the more word-like it is and the faster a lexicality decision can be executed.   We used the correlation $r$ of $\hat{\mathbf{s}}$ and $\mathbf{s}$ as a measure of semantic proximity. Since for large $r$, lexical decision times are expected to be short, whereas for small $r$ long decision times are more likely, we took $1-r$ as a measure that we expect to enter into a positive correlation with RT. This measure, when based on FIL, has a density that is roughly symmetrical, and that does not require further transformations to avoid adverse affects of outliers.

The panels in the middle row present the partial effect of $1-r$ as predictor of RT using FIL (AIC -12,900.75), and the bottom panels present the corresponding partial effects using EL (AIC -10,896.19). The GAM with the FIL-based predictor clearly provides the superior fit to the observed response latencies. The effect of $1-r$ on the mean is fairly linear for FIL but U-shaped for EL.  A strong effect on the variance is present for FIL, but absent for EL.  Effects are opposite to those of frequency, as expected, as $1-r$ is constructed to be positively correlated with RT.
The absence of a highly significant effect on the variance in RT for EL ($p = 0.0356$) is perhaps unsurprising given that EL learns the mapping from form to meaning to perfection (within the constraints of a linear mapping and the type distributions of forms and meanings), and hence little predictivity for human processing variance is to be expected. The question of why $1-r$ has a U-shaped effect on RT for EL will be addressed below.  

In summary, FIL generates a predictor ($1-r$) that is better aligned with observed RTs.  However, log frequency provides a better fit (AIC -16,384.8). Here, it should be kept in mind that word frequency is correlated with many other lexical properties \citep{baayen2010demythologizing}, including word length and number of senses.  Longer, typically less frequent, words require more fixations, and hence are expected to have longer reaction times.  The greater number of senses for higher-frequency words are not directly reflected in the embeddings, which typically consist of a single embedding per unique word form, rendering the mapping less precise. As a consequence, there is necessarily imprecision in measures derived from our learning models.   Furthermore,  $1-r$ is only one of the many learning-based measures that predict lexical decision times, see \citet{heitmeier2023trial} for detailed discussion. 

\begin{figure}[!htb]
\centering
\includegraphics[width=\textwidth]{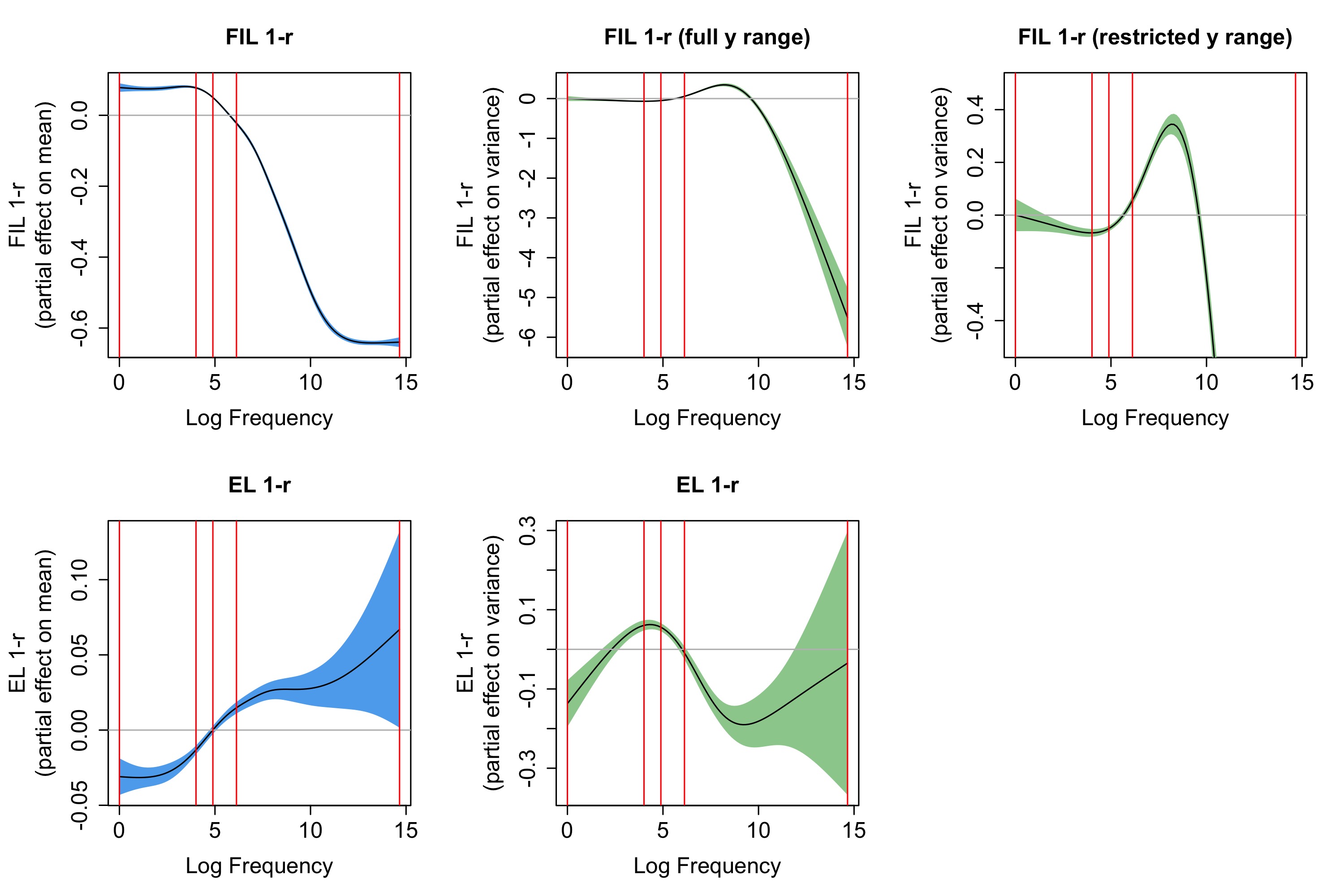}
\caption{
Partial effects for mean (left, confidence intervals in blue) and variance (center and right, confidence intervals in green, the y-axis is on the $\log(\sigma - 0.01)$ scale), for Gaussian Location-Scale GAMs predicting $1-r$ from log frequency for FIL (upper panels), and for EL (bottom panels). The panel in the upper right zooms in on the partial effect of variance shown to its left. The vertical red lines represent the quartiles of log frequency. \\
FIL 1-r shows a similar S-shaped curve as a function of frequency as observed for reaction times, but does not have a similar effect as frequency on the variance in the reaction times.}
\label{fig:freqAndLearn}
\end{figure}

What we have not considered thus far is how $1-r$ is affected by frequency when learning is based on FIL and on EL, and how the effect of frequency on these measures compares to the effect of frequency on reaction times. Figure~\ref{fig:freqAndLearn} presents the partial effects of Gaussian Location-Scale GAMs for FIL (upper panels) and EL (lower panels). The leftmost panels present effects for the mean, and the center panels effects for the variance. For FIL, the highest-frequency words are learned to perfection, and hence the variance in $1-r$ is extremely small.  To highlight the variance function for most of the datapoints, the upper right panel restricts the y-axis to a smaller range. For all points in the interquartile range of frequency, the variance increases with frequency. 

Comparing the partial effects for the means, FIL presents a curve that is similar to the partial effect of log frequency on reaction time (see Figure~\ref{fig:FIL_EL_RT_plot}, upper left panel), whereas the curve for EL is an increasing function of frequency, rather than a decreasing function.   The reason for this is straightforward: higher-frequency words share more segments and substrings with other words than lower frequency words, they are shorter, and tend to have more form neighbors \citep{Nusbaum:85,Baayen:2001}. As a consequence, they provide less information for the mapping from form to meaning, resulting in less accurate predicted semantic vectors, and hence higher values of $1-r$.   This disadvantage of being shorter and easier to pronounce is overcome in FIL. FIL, and also incremental learning, provide  higher frequency words with more learning opportunities compared to lower-frequency words. 

The U-shaped curve of $1-r$ using EL as predictor of reaction time (see the lower left panel of Figure~\ref{fig:FIL_EL_RT_plot}) can now be understood.  For EL, median $1-r$ is 0.32, which is where the curve reaches its minimum.  As we move to the left of the median, word frequency goes down, and as a consequence, salient segments and segment combinations (cf. English \textit{qaid}, `tribal chieftain', which has the highly infrequent bigram \textit{qa}) are more common.  These salient segments and n-grams allow these words to map precisely onto their meanings, much more so than is warranted by their very low frequencies of use. Although EL provides estimates of $1-r$ that are low, EL underestimates the difficulty of learning these words in actual usage.  As a consequence, actual reaction times are higher than expected given the computed $1-r$. Conversely, when we move from the median to the right, we see the expected slowing due to being further away from the semantic target.  Apparently, the greater form similarity and denser form neighborhoods that characterize higher-frequency words results in estimates of $1-r$ that are reasonably aligned with reaction times, albeit by far not as well as when FIL is used to estimate $1-r$.

We have seen that the variance in RTs goes down as word frequency is increased, which we attributed to higher frequency words being known and used by more speakers than is the case for lower-frequency words, and the general reduction in variability that comes with increased practise.  These kinds of factors are not taken into account in the current FIL mapping. It is therefore interesting to see that without these factors, FIL suggests that, at least for the  interquartile range of frequency, the variance increases with frequency.  But why this might be so is unclear to us.  

Considered jointly, these results provide good evidence that FIL adequately integrates frequency into linear discriminative learning, outperforming endstate learning by a wide margin, both qualitatively and quantitatively.

\subsection{Spoken word recognition in Mandarin}\label{sec:mandarin}

Mandarin, as a tone language, alters pitch patterns to distinguish word meanings. There are four lexical tones in Mandarin: high level, rising, dipping, and falling, which will, for convenience, henceforth be referred to as T1, T2, T3, and T4 respectively. Take the syllable \textit{ma} for example. It could mean `mother', `hemp', `horse', or `scorn', depending on which lexical tone it is associated with. 

The role that tone plays in Mandarin word recognition has been widely discussed. Specifically, researchers are interested in whether native listeners exploit tonal information similarly as they do for segmental information. In other words, will a mismatch in tone (e.g., \textit{ma3} and \textit{ma1}) reduce the activation of a given word, to the same extent as a mismatch in segments (e.g., \textit{ba1} and \textit{ma1})? \citet{lee2007does} addresses this issue with an auditory priming experiment. In his study, four priming conditions were designed, as shown in Table \ref{tab:lee_2007}. Among them, ST is an identity priming condition, and UR is a control priming condition. The experimentally critical conditions are S and T, where only either syllable or tone is shared between primes and targets. If tonal information is as important as segmental information in Mandarin, then the degree of priming should be similar for both conditions. On the contrary, differences should be observed if the two sources of information are treated differently by native listeners.

\begin{table}[]
    \centering
    \begin{tabular}{cccc} \hline
    Prime            &  Target       &  Condition   \\ \hline
    \textbf{chun1}   &  \textbf{chun1}   &     ST       \\ 
    \textbf{chun}3   &  \textbf{chun}1   &     S       \\ 
    tong\textbf{1}   &  chun\textbf{1}   &     T       \\ 
    liao2           &   chun1   &     UR       \\\hline
    \end{tabular}
    \caption{The design of priming conditions for Experiments 1 and 2 of \citet{lee2007does}. For the target word \textit{chun1} `spring', the prime either shares both syllable and tone with it (ST), or only syllable (S) or only tone (T). In the UR condition, neither syllable nor tone is the same. The parts shared between primes and targets are marked in bold.}
    \label{tab:lee_2007}
\end{table}

It was found that reaction times to the target words are shortest in the ST condition, hence most priming, as expected. Interestingly, mere tone sharing or syllable sharing is not sufficient to induce a reliable priming effect:  
the T condition does not differ significantly from the UR condition, whereas the S condition differs from UR only in item analysis, but not in subject analysis.
But importantly, there is still a significant difference between the S and T conditions, where syllable sharing induces faster responses than tone sharing. In other words, more priming is found for syllable sharing than tone sharing. It is noteworthy that this pattern of results holds regardless of whether a long (250 ms, experiment 1) or short (50 ms, experiment 2) inter-stimulus interval is adopted in the experiment.

To model this priming experiment, we made use of the Chinese Lexical Database \citep{sun2018chinese}. In total 48,274 one- to four-character Mandarin words were selected, which include all the stimuli of the experiment, and for which fasttext word embeddings \citep{grave2018learning} are available. For cue representations, following \citet{chuang2021bilingual}, we created segmental and suprasegmental (tonal) cues separately. Thus, for a bisyllabic word such as \textit{wen4ti2} `question', the segmental cues will be triphones of \texttt{\#we}, \texttt{wen}, \texttt{ent}, \texttt{nti}, \texttt{ti\#}, and there will also be tritone cues of \texttt{\#42} and \texttt{42\#}. The separation of segmental and suprasegmental information, however, does not do justice to the fact that segments do have influence on tonal realizations and vice versa \citep[e.g.,][]{howie1974domain, ho1976acoustic, xu2001pitch, fon2007positional}. We therefore also made cues out of tone-segment combinations. To operationalize this, we marked tones on vowels, so that vowels with different tones are treated as separate phones. For the word \textit{wen4ti2}, we then have additional tone-segment triphone cues of \texttt{\#we4}, \texttt{we4n}, \texttt{e4nt}, \texttt{nti2}, \texttt{ti2\#}. This resulted in an overall form vector dimensionality of 47,791.

We ran two LDL models, one with EL and the other with FIL. With EL, comprehension accuracy is at 83.71\%. The accuracy is substantially worse with FIL; \texttt{accuracy@1} is 8.98\%. As discussed previously, this is largely due to the low accuracy for especially low frequency words. After taking token frequency into account, the frequency-weighted accuracy is at 86.89\%. 

\begin{figure}
    \centering
    \includegraphics[width=.8\textwidth]{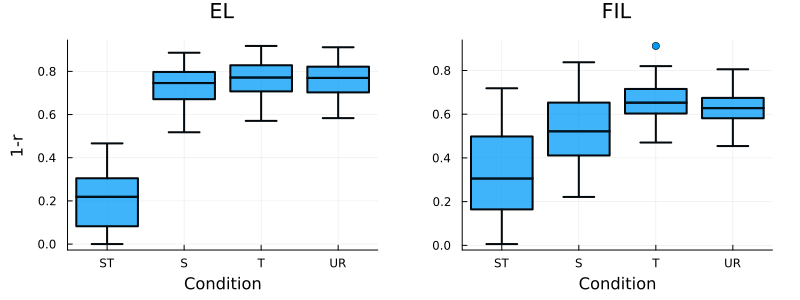}
    \caption{Boxplots of LDL simulated RTs for the four priming conditions in \citet{lee2007does} with EL (left) and FIL (right).\\
    FIL correctly predicts the experimental results of Lee (2007).}
    \label{fig:mand_sim}
\end{figure}

To model the RTs of the auditory priming experiment of \citet{lee2007does}, we calculated the correlation between the predicted semantic vector of the prime word ($\hat{s}_{prime}$) and the gold standard semantic vector of the target word ($s_{target}$), and again took $1 - r$ to predict RTs \citep[see][for the same implementation to simulate RTs for visual priming experiments]{baayen2020modeling}. Results of simulated RTs with EL and FIL are presented in Figure \ref{fig:mand_sim}. For EL (left panel), the simulated RTs of the ST condition is the shortest, unsurprisingly. For both the S and T conditions, the simulated RTs are similar to those of the UR condition, indicative of no priming. Tukey's HSD test reported no significant difference for any pairwise comparison among the S, T, and UR conditions. Although the general pattern of results is in line with the behavioral data, we however missed the crucial difference between the S and T conditions. 

A different picture emerges with the FIL modeling. As shown in the right panel of Figure \ref{fig:mand_sim}, not only does the ST condition induce faster responses than the other three conditions, but the simulated RTs of the S condition are also significantly shorter than those of the T condition ($p < .0001$, according to a Tukey's HSD test), as was found in the behavioral data. 
We note that when compared to the UR condition, the S condition induces significantly shorter simulated RTs, but not the T condition. This pattern of results also to a large extent replicates the empirical findings of \citet{lee2007does}, as the S-UR difference is significant in item analysis but not in subject analysis. As both the S-T and S-UR differences
are absent in EL modeling, we conclude that FIL provides an estimate that better approximates the actual auditory comprehension of native Mandarin listeners.

\section{But what about order?}\label{sec:childes}

Arguably, there is a piece of information missing when using FIL: order information. If a word is highly frequent early in a learning history and never occurs later, would it be forgotten by a WHL model but learned fairly well by a FIL model? To investigate how much of a problem this loss of order is in real-world longitudinal data, we used data of ``Sarah'' in the CHILDES Brown corpus \citep{brown1973first} and of ``Lily'' in the PhonBank English Providence corpus \citep{demuth2006word}. To access all child-directed speech, we utilised the \textit{childesr} package \citep{sanchez2019childes} to extract the data of all utterances not made by the target children themselves, resulting in 189,772 and 373,606 tokens respectively\footnote{We ordered the data by age, followed by utterance id, followed by the token sequence within the utterance. Unfortunately, this part of \textit{childesr} is not very well documented, so though sampling suggests that this corresponds to the original order of the data, we cannot be absolutely sure. However, the overall ordering by age should ensure that our results are valid even if some of the utterances might be in the wrong order.}. Of these, we kept all tokens for which pronunciations were available in CELEX \citep{baayen1995celex} and for which we could obtain word embeddings in 300-dimensional Wikipedia2Vec word embeddings \citep{yamada2020wikipedia2vec}. This resulted in 162,443 learning events (3,865 unique word tokens) for Sarah and 326,518 learning events (7,433 unique words tokens) for Lily. The cue matrix was created based on bigrams of CELEX DISC notation symbols, so for example, \textit{thing} was represented as \texttt{\#T}, \texttt{TI}, \texttt{IN}, \texttt{N\#}, in order to model auditory comprehension (thus, the form vector dimensionality was 943 and 1,087 for Sarah and Lily respectively). We then trained the comprehension matrix $\mathbf{F}$ incrementally, evaluating after every 5,000 learning events the correlation of the predicted semantics of all word tokens with their target semantics, as well as keeping track of the frequency of each word token within the last 5,000 learning events. To gauge the effect of different learning rates we ran this simulation for $\eta \in \{0.1, 0.01, 0.001, 0.0001\}$.

\begin{table}[!htb]
    \centering
    \begin{tabular}{l| l|c|c|c|c}
    \hline
        method & $\eta$ & \multicolumn{2}{c|}{correlation \texttt{accuracy@1}} & \multicolumn{2}{c}{$r$(WHL, FIL)} \\
        & & Sarah & Lily & Sarah & Lily \\
        \hline
        \hline
        FIL & ---& 12.9\% & 8.1\% & --- & ---\\
        \hline
        \multirow{4}{*}{WHL} & 0.1 & \textbf{22.3\%}  & \textbf{13.8\%} & .79  & .73 \\
        &0.01 & 15.7\% & 10.9\% & \textbf{.97} & .94 \\
        &0.001 & 6.9\%  & 5.4\% & .95  & \textbf{.97} \\
        &0.0001 & 2.1\% & 1.5\% & .78 & .80 \\
        \hline
    \end{tabular}
    \caption{Correlation \texttt{accuracies@1} for FIL and WHL with different learning rates ($\eta$). $r$(WHL, FIL) indicates the correlation between the target correlations obtained with WHL and the target correlations obtained with FIL (see Figure~\ref{fig:correlations} for scatter plots). ``Sarah'' and ``Lily'' correspond to the child-directed speech in the Brown corpus \citep{brown1973first} and Providence corpus \citep{demuth2006word} respectively.}
    \label{tab:whl_fil_accs}
\end{table}

\begin{figure}[!htb]
     \centering
    \includegraphics[width=\textwidth]{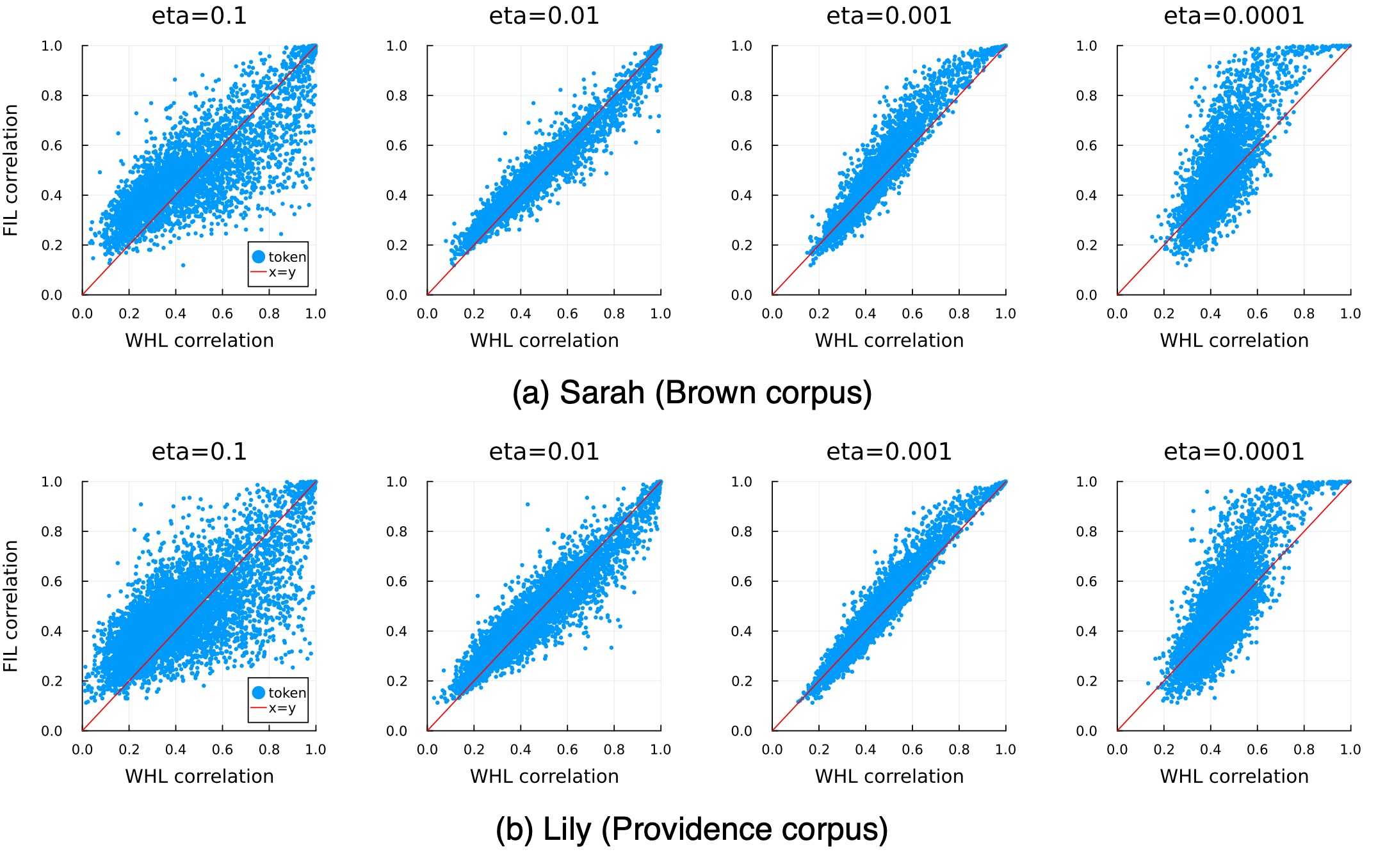}
    \caption{Correlation of WHL learned predicted semantics with their targets against correlations of FIL learned predicted semantics with their targets (blue dots), for different learning rates, and for Sarah (a) and Lily (b). The diagonal red lines denote the x=y line. \\
    WHL target correlations and FIL target correlations are highly correlated, with the tightest relationship visible for $\eta = 0.01$ for Sarah and $\eta=0.001$ for Lily (see also Table~\ref{tab:whl_fil_accs}).}
    \label{fig:correlations}
\end{figure}

In order to investigate the consequences of neglecting order during learning, we also trained a model with the same form and semantic matrix but using the FIL method. The FIL method results in a correlation accuracy of 12.9\% for Sarah and 8.1\% for Lily. For WHL the correlation accuracies vary across learning rates (Table~\ref{tab:whl_fil_accs}), with accuracy decreasing for lower learning rates. FIL correlation accuracies are somewhat better than WHL accuracies for $\eta=0.001$ and somewhat worse than for $\eta=0.01$. Target correlations obtained with FIL and WHL are in general remarkably similar, correlated the highest for learning rates of $0.001$ and $0.01$ for ``Lily'' and ``Sarah'' respectively. This can also be observed visually in Figure~\ref{fig:correlations}: WHL and FIL target correlations are the least similar for $\eta=0.1$ and $\eta=0.0001$. Interestingly, for higher learning rates, low WHL correlations tend to be higher in FIL and vice versa, whereas for lower learning rates an advantage of FIL over WHL is more visible for higher accuracies, while low accuracies in WHL tend to be even lower in FIL.

\begin{figure}[!htb]
     \centering
\includegraphics[width=.8\textwidth]{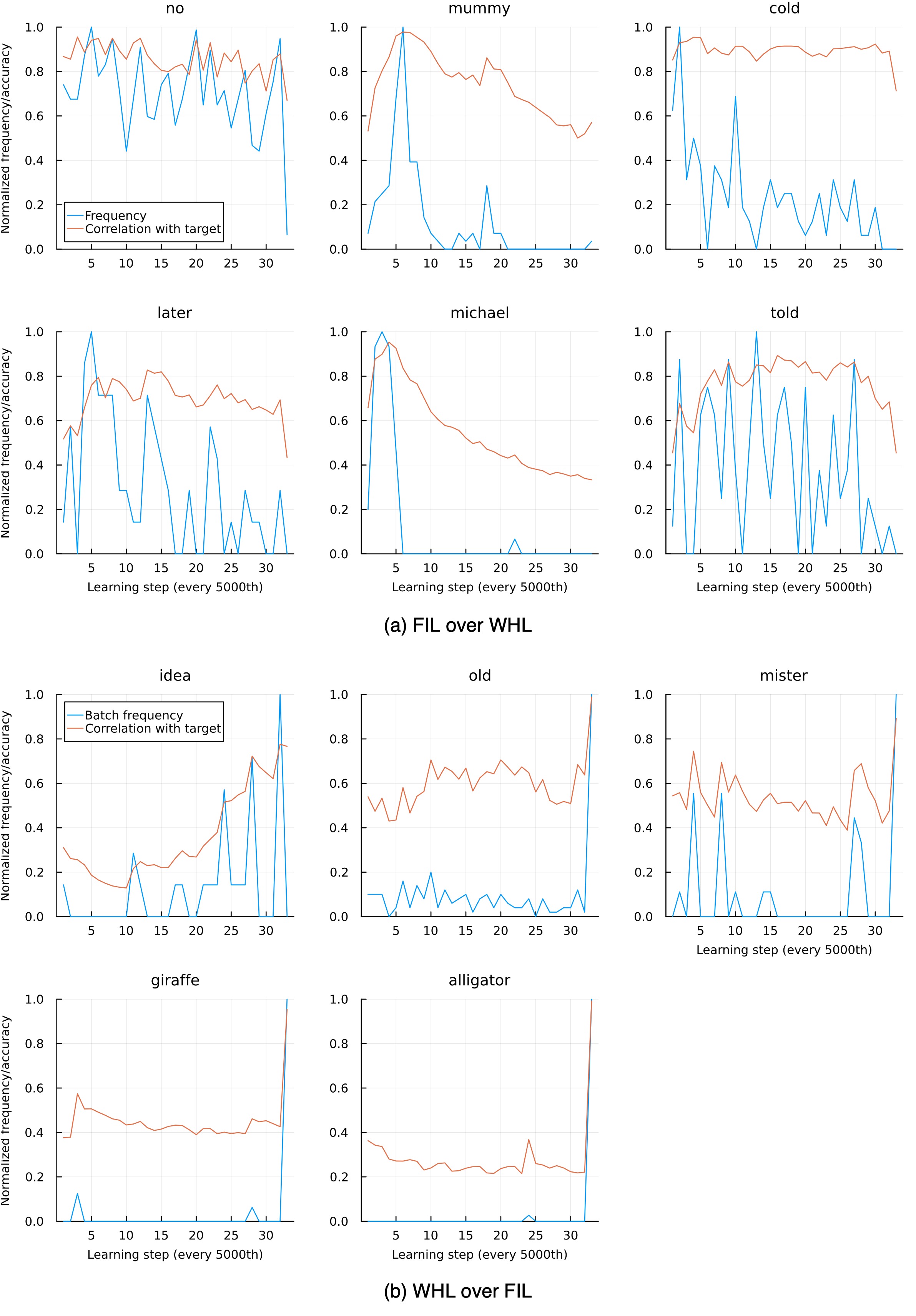}
        \caption{Individual tokens where FIL and WHL correlations to target differ clearly, taken from ``Sarah'', trained with WHL and $\eta=0.01$. (a) shows cases where FIL outperforms WHL, (b) where WHL outperforms FIL. Frequencies are normalised by their maximal frequency inside a learning batch of 5000 learning events.\\
        In (a), frequency tends to decrease over time. Thus, the item is learned well in the beginning, and the correlation with target goes down over time. In (b)} the opposite is the case.\label{fig:outliers}
\end{figure}

Striking a balance between WHL accuracy and correlation between WHL and FIL accuracies, we now focus on $\eta=0.01$. As can be seen in Figure \ref{fig:correlations} there are a few outliers where either FIL clearly outperforms WHL or vice versa. For the former case, the most apparent cases for Sarah are \textit{no}, \textit{mummy}, \textit{cold}, \textit{later}, \textit{michael} and \textit{told}. Except \textit{no} which suffers due to its homophone \textit{know}, all show a relatively higher frequency at the beginning of the learning trajectory compared to the end (see upper panel of Figure~\ref{fig:outliers}). Moreover, they tend to have overlapping di-phones: for example, \textit{told} has overlap with \textit{old} with which they are confused in the WHL model. Lily's data shows a similar pattern. WHL therefore unlearns when two conditions apply: a word occurs very infrequently in the later stages of the learning sequence and it has overlaps with cues in other words. WHL has an advantage over FIL when words have a higher frequency later in the sequence: for Sarah's data this is the case for \textit{idea}, \textit{old}, \textit{mister}, \textit{giraffe} and \textit{alligator} (see lower panel of Figure~\ref{fig:outliers}). 

\begin{figure}[!htb]
    \centering
     \includegraphics[width=.9\textwidth]{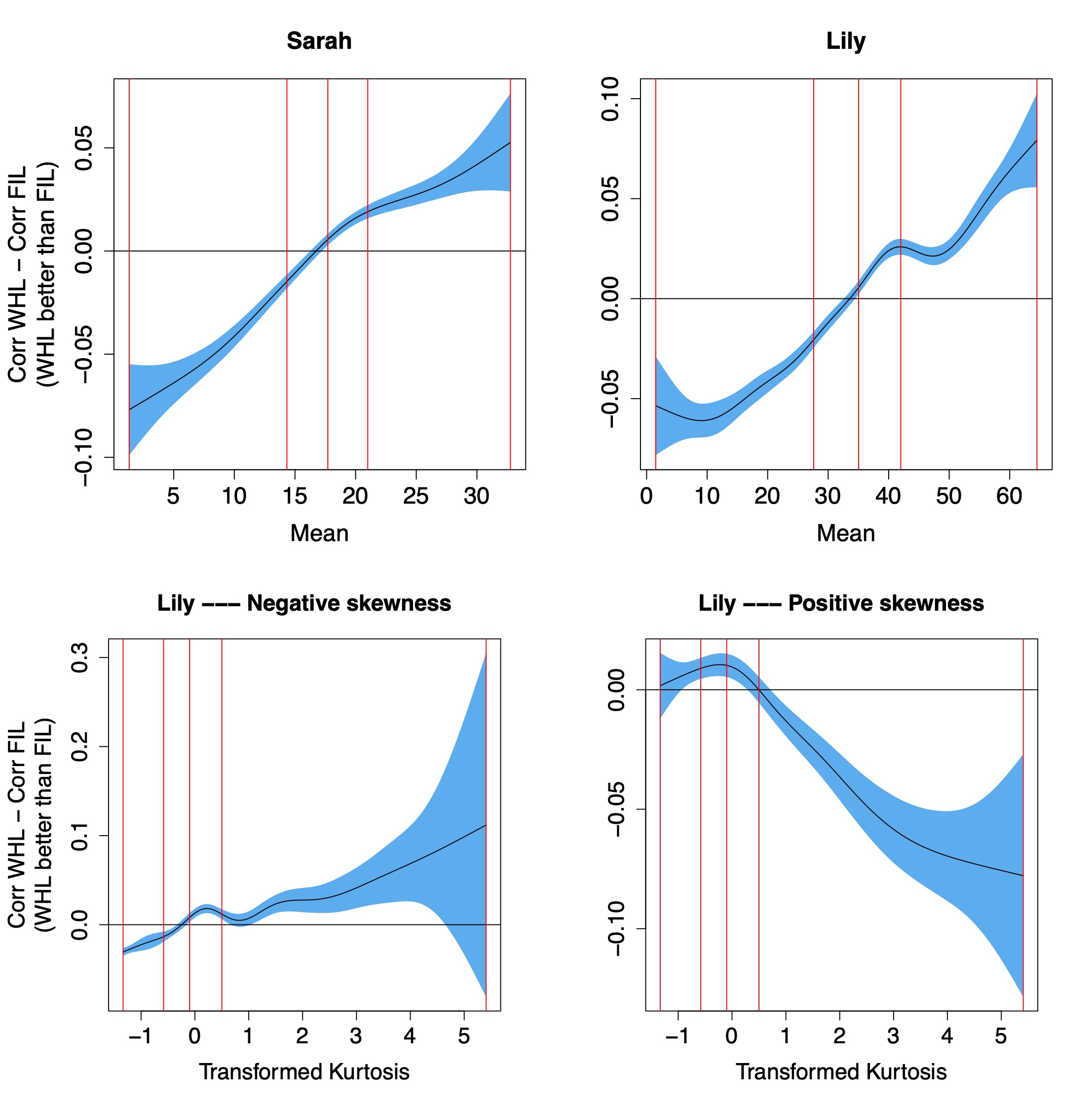}
        \caption{Predicting the difference between target correlations in WHL and in FIL from frequency distribution of words across time. The y-axes show partial effects. \textbf{Top row:} The higher the mean (i.e. higher frequencies at later time steps, see ``giraffe'' and ``alligator'' in the lower row of Figure~\ref{fig:outliers} for examples of words with a high mean), the better WHL performs than FIL. The vertical red lines represent the 0\%, 25\%, 50\%, 75\%, and 100\% percentiles. \textbf{Bottom row:} For negative skew (higher frequencies at later time steps), the peakier the distribution (higher kurtosis), the larger the advantage of WHL over FIL, and vice versa for positive skew. Kurtosis was transformed to reduce outlier effects, details in Supplementary Materials.\\
        WHL outperforms FIL for words with higher frequencies at later learning steps.}
        \label{fig:mean}
\end{figure}

To confirm this qualitative analysis quantitatively, we computed the mean, mode, skewness and kurtosis of the frequency distribution across time for each word. We then used these as predictors in individual (due to collinearity) Generalised Additive Models \citep[GAMs;][]{wood2011gam} to predict the difference in target correlation in WHL and  FIL for $\eta=0.01$, i.e. predicting whether WHL or FIL perform better depending on the frequency distribution of a word across time (details on method in Supplementary Materials). We found that the higher the mean and mode, and the more negative the skew of the frequency distribution (all indicating that frequencies are higher at later timesteps; compare e.g. ``giraffe'' and ``alligator'' in Figure~\ref{fig:outliers} for examples of high mean and mode and low skewness, with ``mummy'' and ``michael'' in Figure~\ref{fig:outliers} for examples of low mean and mode and high skewness), the better WHL performed than FIL. To illustrate, we show the effect of the mean on performance in the upper row of Figure~\ref{fig:mean}. Additionally, when kurtosis was entered (differentiated by whether skewness was positive or negative), an interesting effect emerged: for negative skewness values (high frequency at later time steps), more positive kurtosis (peakier distribution) yielded an advantage of WHL, while a peakier distribution of positive skewness values led to an advantage of FIL (see bottom row of Figure~\ref{fig:mean}). All these results confirmed our qualitative analysis above. 

To summarise, some interesting phenomena related to order information get lost when using FIL. For higher learning rates, WHL overall yields higher accuracies, possibly because it is sensitive to the burstiness of frequency during the course of learning. On the other hand, for words that are equally distributed across learning events, the predictions of WHL and FIL are very similar, and the two methods thus result in highly similar correlations with their respective target semantics \citep[see also][]{milin2020keeping}.

\section{Discussion}

We have introduced a new way for estimating mappings between form and meaning in the Discriminative Lexicon Model \citep[DLM;][]{baayen2018inflectional, baayen2019discriminative} that takes frequency of use into account, Frequency-Informed Learning (FIL), complementing incremental learning with the learning rule of Widrow-Hoff (WHL) and endstate learning using  multivariate multiple regression (EL).  Each of these methods has advantages as well as disadvantages.

\subsection{Three methods for computing mappings in the DLM}

WHL enables trial to trial learning and hence is, in principle, an excellent choice for datasets with learning events that are ordered in time. Examples of such datasets are child-directed speech in the CHILDES database ordered by the age of the addressee and the time-series of reaction times in mega-experiments \citep{heitmeier2023trial}.  The disadvantage of WHL is that it is computationally demanding, and prohibitively so for large datasets.

FIL offers a computationally lean way of taking frequency of use into account, but it is insensitive to the order of learning events. It is therefore an excellent choice for datasets with learning events that are unordered, which is typically the case for data compiled from corpora or databases.  For large datasets with temporally ordered learning events, FIL can be applied to a sequence of datasets with increasing sample sizes to probe how learning develops over time. How exactly such sequential modeling compares with WHL is a topic for further research. 

Models using EL are not computationally demanding, but they are also not sensitive to the frequencies with which learning events occur.  For usage-based approaches to language \citep[see, e.g.,][]{Bybee:2010}, this is a serious drawback.  Nevertheless, EL has an important advantage of its own: it provides a window on what can be learned in principle, with infinite experience. In other words, EL is a good analytical tool for any datasets for which a type-based analysis is appropriate or insightful. For instance, if one's interest is in how well Dutch final devoicing can be mastered on the basis of subliminal learning only, the EL model informs us that a comprehension accuracy of 83\% can be reached \citep[see also][]{heitmeier2023linear}.  When measures are gleaned from an EL model and used as predictors for aspects of lexical processing, it will typically be necessary to include a measure of frequency of use as a covariate.

With FIL, however, we have an analytical instrument that integrates experience into mappings between form and meaning. It obviates the practical necessity, when using WHL, of scaling frequencies down, nor is a log-transform of usage required. The latter is particularly undesirable as it artefactually boosts performance for low-frequency words while degrading performance for high-frequency words.

An open question when it comes to comparing the three methods is whether measures of the models' accuracy should be type-based, i.e. all words contribute equally to the overall accuracy measure, or whether it should be token-based. Previous models have usually been evaluated based on type-accuracy, but a high token-accuracy in word recognition might be of more practical use in every day life than a high type- but low token-accuracy, where a model is able to recognise many low frequency words but struggles with many of the high frequency words with which speakers are constantly confronted. In this context, it is worth keeping in mind the forgetting curve of Ebbinghaus \citep{ebbinghaus1885gedachtnis} and recent advanced methods for repeating  words at exactly the right moment in time to optimize fact learning, including vocabulary learning \citep{vanRijn2022capturing}.  From this literature, it is crystal clear that words encountered only once, which in corpora typically constitute roughly 50\% of word types, cannot be learned with a single exposure.   Whereas EL does not capture effects of frequency, FIL clarifies that once we start taking frequency into account, it is practice that makes perfect.

\subsection{The relationship between word frequency and lexical decision reaction times}

A surprising property of FIL is that the correlation $r$ of the predicted semantic vector with its gold standard target semantic vector emerges as a key to understanding two findings in lexical decision tasks. We found that FIL is crucial for modelling a stronger priming effect of segmental information compared to tone information in an auditory lexical decision task in Mandarin Chinese.  For unprimed lexical decisions as available in the Dutch Lexicon Project,
reasoning that greater correlation should afford shorter decision times, we used $1-r$ as an approximation of simulated decision times. We found that 
$1-r$ based on FIL provided much improved prediction accuracy compared to $1-r$ based on EL. 
Moreover, FIL also provides insights into the non-linear nature of the word frequency effect in the lexical decision task.   We showed that the mirror-sigmoid relation between empirical decision times and log frequency emerging from a GAM analysis also characterizes the functional relation between `simulated decision times' $1-r$ and log frequency.  This suggests that FIL successfully filters usage through discriminative learning to obtain estimates of how well the meanings of words are understood.

This finding fits in well with a recent debate that was re-sparked by \citet{murray2004serial} arguing that rank-transformed frequencies account for lexical decision reaction times better than log-transformed ones and that, therefore, serial-search models should not be discounted as models of word recognition.\footnote{However, note that the variance of word frequencies is similar in magnitude to the frequency itself (under the assumption that word frequencies are Poisson-distributed).  By moving from frequencies to ranks, differences in frequency that seem large but that will vary substantially across samples are reduced to much smaller differences in ranks.  In the light of these considerations, the excellent predictivity of rank for reaction times is due to distributional properties of the language in combination with sampling error, rather than due to serial searches in frequency ordered mental word lists.
}
Recently, \citet{kapatsinski2022logistic} showed that log frequencies transformed by the logistic function (a function frequently used in deep learning models) predict reaction times in the same way as a rank-transformation, implying that the linear relationship between rank frequency and reaction times is not necessarily evidence in favour of the serial search model. Since the DLM is not a classification model, we do not make use of the logistic function but use correlation to compute how close a word's predicted meaning is to its true meaning. The estimated functional relation between -1000/RT and $1-r$ estimated with FIL is close to linear (see Figure~\ref{fig:FIL_EL_RT_plot}, panel (2,1)), and a very similar partial effect emerges when the untransformed RTs are regressed on $1-r$. Thus, a linear relationship between a FIL-based predictor, $1-r$ (or equivalently, $r$) and reaction times falls out directly of the DLM, without requiring further transformations. This provides further evidence in favour of theories suggesting that frequency effects arise due to the distributional properties  of words' forms and meanings during learning. \footnote{See \citet{norris2006bayesian} for a similar argument in the context of a Bayesian approach to the role of word frequency in lexical processing.}

A question that we leave to  future research is whether measures derived from FIL mappings will obviate the need to include frequency of use as a covariate,  reduce the variable importance of this predictor, or complement it. One possible complication here is that while frequency-informed mappings make the DLM predictions more realistic, they naturally also create a confound with word frequency, which needs to be teased apart from other (e.g. semantic) effects captured by the DLM \citep[see][for further discussion]{Gahl2022thyme}. Furthermore, in the present study we did not take into account (any measure of) a word's context \citep[e.g.][]{adelman2008modeling, hollis2020delineating}. \citet{Gahl2022thyme} explore an additional measure of frequency which is informed by a word's context by measuring how strongly a word predicts itself in any utterances it occurs in. This measure outperforms frequency for predicting acoustic durations of English homophones, and it also outperforms frequency as a predictor of visual lexical decision latencies. This measure may complement the measures from the DLM.

What complicates such comparisons is that frequencies based on corpora typically concern lexical use in a language community, whereas individual speakers have very different experiences with lower-frequency words, depending on their interests, professions, education, and reading/speaking habits \citep[e.g.][]{kuperman2013reassessing, baayen2016frequency}. For instance, about 5\% of the Dutch words in the present dataset are unknown to the fourth author. These are low-frequency words that the EL gets correct in 30 out of 35 cases. By contrast, the FIL gets only 2 out of these 35 cases correct. In a similar vein, \citet{diependaele2012noisy} showed that in the DLP, when reacting twice to the same lowest-frequency word stimuli in the DLP participants only agree in about 50\% of cases, bringing their accuracy to chance level. 

The inter- and intra-variability of subjects' word knowledge might also be a reason why FIL underestimates the variance of reaction times in the Dutch Lexicon Project for low-frequency words. Naturally, FIL, being based on `community' frequencies, is not able to account for this effect. Our results, however, highlight the importance of modelling not only the mean but also the variance of reaction times, as it might provide a further window into speakers' word recognition process and differentiating between different models attempting to account for the observed behavioural data. To our knowledge, previous models of word recognition have not attempted to account for both the mean and variance of reaction times \citep[but see][for a model attempting to also predict the distribution of reaction times]{ratcliff2004diffusion}.

When evaluating the accuracy of a FIL model, having the closest proximity to the semantic vector of the target word is probably too stringent a criterion.  Especially for lower-frequency words, having a rank among the top $k$ (for low $k$)  nearest neighbors may be a more reasonable criterion.  The reason is that for lower frequency words, knowledge of what words may mean is only approximate.  For instance, although the names of gemstones such as jasper, ruby, tourmaline, and beryl will be known to many readers, picking out the ruby from four reddish pieces of rock requires more precise knowledge than is available to the authors.  Especially for lexical decision making, being `close enough' may be good enough for a yes-response \citep[see][for detailed discussion of lexical decision making]{heitmeier2023trial}.

An additional complication is that low-frequency words often have multiple, equally rare, meanings. (By contrast, for high-frequency words, one often finds that of a set of possible meanings, one is dominant.) By way of example, the low-frequency Dutch word \textit{bras} in our dataset can denote `uncooked but peeled rice', `junk', and `a specific set of ropes used on sailing ships'.  This may provide further explanation of why the correlation measure ($1-r$) underestimates the variance in decision latencies for lower-frequency words as compared to higher-frequency words.  

\subsection{Computational modelling of word frequency effects}

Our findings are particularly enlightening when comparing them to accounts of word frequency effects in previous models of word recognition. To answer the three key questions regarding the word frequency effects identified in the introduction, in the DLM, frequency effects arise as a consequence of the input distribution it is trained on. Similarly to other network models, the effect of word frequency is stored in the model's mapping weights. The DLM is an extremely simple computational model involving only linear mappings, which can be efficiently numerically computed. As demonstrated, the resulting mappings are very similar to those obtained by training the model incrementally on unordered data \citep[as is done in all deep learning models utilising backpropagation, such as the triangle model,][]{seidenberg1989distributed}. As such, it enables the modeller to answer many questions related to the word frequency effect, such as that of individual differences, in a computationally lean way. Finally, we demonstrate that the DLM is able to account for the non-linear relationship between frequency and reaction times in lexical decision without requiring to use a log-transformation \citep[cf.][]{mcclelland1981interactive, rumelhart1982interactive, seidenberg1989distributed}. Moreover, the DLM can not only account for frequency effects but also enables investigation of the influence of words' more finegrained characteristics on reaction times. Crucially, these are not independent. The DLM's predicted reaction times  also depend on words' form and meaning properties, not only their frequency distribution. 

Therefore, it is important to note that the DLM's performance also depends on many modelling choices, such as the chosen form granularity, semantic vectors etc. Ideal modelling choices can often vary across languages --- for instance, while for English, Dutch and German, trigrams are often the unit of choice \citep{heitmeier2023trial, heitmeier2021modeling}, previous work has found that for Vietnamese, bigrams are preferable \citep{pham2015vietnamese}, while for Maltese, Kinyarwanda and Korean, form representations based on syllables perform well (
\citeauthor{nieder2023discriminative}, \citeyear{nieder2023discriminative}; \citeauthor{van2023comprehension}, \citeyear{van2023comprehension}, \citeauthor{chuang2021vector}, \citeyear{chuang2021vector}; an in-depth discussion of the various considerations when modelling a language with the DLM can be found in \citeauthor{heitmeier2021modeling}, \citeyear{heitmeier2021modeling}). While the present study is limited to Dutch, Mandarin and English, future work should further verify the efficacy of FIL  on morphologically more diverse languages.

To put the present results in a broader perspective, consider the characterization given by \citet{breiman2001statistical} of machine learning on the one hand and statistics on the other hand.  They argue that machine learning aims at obtaining accurate predictions. How these predictions are obtained and why a technique works is not of interest.  By contrast, statistics aims to formulate a model that could have generated the data. Within this characterisation --- which may be too extreme \citep[e.g.][]{shmueli2010explain} ---  FIL is much closer to statistical modeling than to machine learning, and it is surprising to us how much can be achieved simply with a form of weighted multivariate multiple regression.  We hope that FIL will prove to be a useful tool not only for modeling data with `community' frequencies but also for exploring, by means of simulation, what the consequences are of individual differences in usage for lexical processing.

\section*{Acknowledgements}
Funded by the Deutsche Forschungsgemeinschaft (DFG, German Research Foundation) under Germany’s Excellence Strategy – EXC number 2064/1 – Project number 390727645, and by the European Research Council,  project WIDE-742545. The authors thank members of the Quantitative Linguistics group for helpful feedback and discussions, and Yu-Hsiang Tseng for helpful comments on an earlier version of this manuscript.

\bibliography{bib_unique}

\begin{thebibliography}{}

\bibitem[Adelman and Brown, 2008]{adelman2008modeling}
Adelman, J.~S. and Brown, G.~D. (2008).
\newblock Modeling lexical decision: the form of frequency and diversity
  effects.
\newblock {\em Psychological Review}, 115(1):214.

\bibitem[Baayen, 2001]{Baayen:2001}
Baayen, R.~H. (2001).
\newblock {\em Word Frequency Distributions}.
\newblock Kluwer Academic Publishers, Dordrecht.

\bibitem[Baayen, 2005]{Baayen:2005}
Baayen, R.~H. (2005).
\newblock Data mining at the intersection of psychology and linguistics.
\newblock In Cutler, A., editor, {\em Twenty-first century psycholinguistics:
  Four cornerstones}, pages 69--83. Erlbaum, Hillsdale, New Jersey.

\bibitem[Baayen, 2010]{baayen2010demythologizing}
Baayen, R.~H. (2010).
\newblock Demythologizing the word frequency effect: A discriminative learning
  perspective.
\newblock {\em The Mental Lexicon}, 5(3):436--461.

\bibitem[Baayen et~al., 2018a]{baayen2018inflectional}
Baayen, R.~H., Chuang, Y.-Y., and Blevins, J.~P. (2018a).
\newblock Inflectional morphology with linear mappings.
\newblock {\em The Mental Lexicon}, 13(2):230--268.

\bibitem[Baayen et~al., 2018b]{Baayen2018a}
Baayen, R.~H., Chuang, Y.-Y., and Heitmeier, M. (2018b).
\newblock {WpmWithLdl}: Implementation of word and paradigm morphology with
  linear discriminative learning.
\newblock R package Version 1.2.20.

\bibitem[Baayen et~al., 2019]{baayen2019discriminative}
Baayen, R.~H., Chuang, Y.-Y., Shafaei-Bajestan, E., and Blevins, J. (2019).
\newblock {The discriminative lexicon: A unified computational model for the
  lexicon and lexical processing in comprehension and production grounded not
  in (de) composition but in linear discriminative learning}.
\newblock {\em Complexity}, 2019.

\bibitem[Baayen et~al., 1997]{Baayen:Dijkstra:Schreuder:1997}
Baayen, R.~H., Dijkstra, T., and Schreuder, R. (1997).
\newblock Singulars and plurals in {D}utch: Evidence for a parallel dual route
  model.
\newblock {\em Journal of {M}emory and {L}anguage}, 36:94--117.

\bibitem[Baayen et~al., 2016]{baayen2016frequency}
Baayen, R.~H., Milin, P., and Ramscar, M. (2016).
\newblock Frequency in lexical processing.
\newblock {\em Aphasiology}, 30(11):1174--1220.

\bibitem[Baayen et~al., 2011]{baayen2011amorphous}
Baayen, R.~H., Milin, P., {\DJ}ur{\dj}evi{\'c}, D.~F., Hendrix, P., and
  Marelli, M. (2011).
\newblock An amorphous model for morphological processing in visual
  comprehension based on naive discriminative learning.
\newblock {\em Psychological review}, 118(3):438.

\bibitem[Baayen et~al., 1995]{baayen1995celex}
Baayen, R.~H., Piepenbrock, R., and Gulikers, L. (1995).
\newblock The {CELEX} lexical database [cd rom].
\newblock Philadelphia: Linguistic Data Consortium, University of Pennsylvania.

\bibitem[Baayen and Smolka, 2020]{baayen2020modeling}
Baayen, R.~H. and Smolka, E. (2020).
\newblock Modeling morphological priming in german with naive discriminative
  learning.
\newblock {\em Frontiers in Communication}, 5:17.

\bibitem[Balota et~al., 2004]{balota2004visual}
Balota, D.~A., Cortese, M.~J., Sergent-Marshall, S.~D., Spieler, D.~H., and
  Yap, M.~J. (2004).
\newblock Visual word recognition of single-syllable words.
\newblock {\em Journal of experimental psychology: General}, 133(2):283.

\bibitem[Beaumont, 1965]{Beaumont:1965}
Beaumont, R.~A. (1965).
\newblock {\em Linear Algebra}.
\newblock Harcourt, New York.

\bibitem[Bezanson et~al., 2017]{bezanson2017julia}
Bezanson, J., Edelman, A., Karpinski, S., and Shah, V.~B. (2017).
\newblock Julia: A fresh approach to numerical computing.
\newblock {\em SIAM review}, 59(1):65--98.

\bibitem[Bird, 2006]{bird2006nltk}
Bird, S. (2006).
\newblock {NLTK: the natural language toolkit}.
\newblock In {\em Proceedings of the COLING/ACL 2006 Interactive Presentation
  Sessions}, pages 69--72, Philadelphia, PA.

\bibitem[Breiman, 2001]{breiman2001statistical}
Breiman, L. (2001).
\newblock Statistical modeling: The two cultures (with comments and a rejoinder
  by the author).
\newblock {\em Statistical science}, 16(3):199--231.

\bibitem[Brown, 1973]{brown1973first}
Brown, R. (1973).
\newblock A first language: The early stages.
\newblock {\em Harvard University}.

\bibitem[Brysbaert et~al., 2011]{brysbaert2011word}
Brysbaert, M., Buchmeier, M., Conrad, M., Jacobs, A.~M., B{\"o}lte, J., and
  B{\"o}hl, A. (2011).
\newblock The word frequency effect.
\newblock {\em Experimental psychology}, 58(5):412--424.

\bibitem[Brysbaert et~al., 2018]{brysbaert2018word}
Brysbaert, M., Mandera, P., and Keuleers, E. (2018).
\newblock The word frequency effect in word processing: An updated review.
\newblock {\em Current Directions in Psychological Science}, 27(1):45--50.

\bibitem[Brysbaert and New, 2009]{brysbaert2009moving}
Brysbaert, M. and New, B. (2009).
\newblock Moving beyond ku{\v{c}}era and francis: A critical evaluation of
  current word frequency norms and the introduction of a new and improved word
  frequency measure for american english.
\newblock {\em Behavior research methods}, 41(4):977--990.

\bibitem[Brysbaert et~al., 2016]{brysbaert2016impact}
Brysbaert, M., Stevens, M., Mandera, P., and Keuleers, E. (2016).
\newblock The impact of word prevalence on lexical decision times: Evidence
  from the dutch lexicon project 2.
\newblock {\em Journal of Experimental Psychology: Human Perception and
  Performance}, 42(3):441.

\bibitem[Bybee, 2010]{Bybee:2010}
Bybee, J. (2010).
\newblock {\em Language, usage and cognition}.
\newblock Cambridge University Press, Cambridge.

\bibitem[Bybee and Hopper, 2001]{bybee2001frequency}
Bybee, J.~L. and Hopper, P.~J. (2001).
\newblock {\em {Frequency and the Emergence of Linguistic Structure}},
  volume~45.
\newblock John Benjamins Publishing, Amsterdam.

\bibitem[Chuang et~al., 2021]{chuang2021bilingual}
Chuang, Y.-Y., Bell, M.~J., Banke, I., and Baayen, R.~H. (2021).
\newblock Bilingual and multilingual mental lexicon: a modeling study with
  linear discriminative learning.
\newblock {\em Language Learning}, 71(S1):219--292.

\bibitem[Chuang et~al., 2022]{chuang2021vector}
Chuang, Y.-Y., Kang, M., Luo, X., and Baayen, R.~H. (2022).
\newblock Vector space morphology with linear discriminative learning.
\newblock In Crepaldi, D., editor, {\em Linguistic morphology in the mind and
  brain.} Routledge, Abingdon.

\bibitem[Chuang et~al., 2020]{chuang2020estonian}
Chuang, Y.-Y., L{\~o}o, K., Blevins, J.~P., and Baayen, R.~H. (2020).
\newblock Estonian case inflection made simple a case study in word and
  paradigm morphology with linear discriminative learning.
\newblock In Körtvélyessy, L. and Štekauer, P., editors, {\em Complex Words:
  Advances in Morphology}, chapter~7, pages 119--14. Cambridge University
  Press.

\bibitem[Demuth et~al., 2006]{demuth2006word}
Demuth, K., Culbertson, J., and Alter, J. (2006).
\newblock Word-minimality, epenthesis and coda licensing in the early
  acquisition of english.
\newblock {\em Language and speech}, 49(2):137--173.

\bibitem[Denistia and Baayen, 2021]{denistia2021morphology}
Denistia, K. and Baayen, R.~H. (2021).
\newblock The morphology of indonesian: Data and quantitative modeling.
\newblock {\em The Routledge Handbook of Asian Linguistics}.

\bibitem[Diependaele et~al., 2012]{diependaele2012noisy}
Diependaele, K., Brysbaert, M., and Neri, P. (2012).
\newblock How noisy is lexical decision?
\newblock {\em Frontiers in psychology}, 3:348.

\bibitem[Ebbinghaus, 1885]{ebbinghaus1885gedachtnis}
Ebbinghaus, H. (1885).
\newblock {\em {\"U}ber das {Gedächtnis}}.
\newblock Dunker and Humbolt, Leipzig.

\bibitem[Ernestus and Baayen, 2003]{ernestus2003predicting}
Ernestus, M. T.~C. and Baayen, R.~H. (2003).
\newblock Predicting the unpredictable: Interpreting neutralized segments in
  dutch.
\newblock {\em Language}, 79(1):5--38.

\bibitem[Faraway, 2005]{Faraway:2005}
Faraway, J.~J. (2005).
\newblock {\em Linear Models with R}.
\newblock Chapman \& Hall/CRC, Boca Raton, FL.

\bibitem[Ferrand et~al., 2010]{ferrand2010french}
Ferrand, L., New, B., Brysbaert, M., Keuleers, E., Bonin, P., M{\'e}ot, A.,
  Augustinova, M., and Pallier, C. (2010).
\newblock The french lexicon project: Lexical decision data for 38,840 french
  words and 38,840 pseudowords.
\newblock {\em Behavior research methods}, 42:488--496.

\bibitem[Fon and Hsu, 2007]{fon2007positional}
Fon, J. and Hsu, H.-j. (2007).
\newblock {Positional and phonotactic effects on the realization of dipping
  tones in Taiwan Mandarin}.
\newblock In Gussenhoven, C. and Riad, T., editors, {\em {Phonology and
  Phonetics, Tones and Tunes: Vol. 2. Experimental Studies in Word and Sentence
  Prosody}}, pages 239--269. Mouton de Gruyter, Berlin.

\bibitem[Forster, 1976]{forster1976accessing}
Forster, K. (1976).
\newblock Accessing the mental lexicon.
\newblock In Wales, F. and Walker, E., editors, {\em New approaches to language
  mechanisms}, pages 257--286. North-Holland, Amsterdam.

\bibitem[Forster, 1979]{forster1979levels}
Forster, K.~I. (1979).
\newblock Levels of processing and the structure of the language processor.
\newblock In {\em Sentence Processing: Psycholinguistic essays presented to
  Merrill Garrett}, Hillsdale, N.J. Erlbaum.

\bibitem[Forster, 1994]{forster1994computational}
Forster, K.~I. (1994).
\newblock Computational modeling and elementary process analysis in visual word
  recognition.
\newblock {\em Journal of Experimental Psychology: Human Perception and
  Performance}, 20(6):1292.

\bibitem[Gahl and Baayen, 2023]{Gahl2022thyme}
Gahl, S. and Baayen, R.~H. (2023).
\newblock Time and thyme again: Connecting spoken word duration to models of
  the mental lexicon.
\newblock {\em Under revision for Language}.

\bibitem[Grave et~al., 2018]{grave2018learning}
Grave, {\'E}., Bojanowski, P., Gupta, P., Joulin, A., and Mikolov, T. (2018).
\newblock Learning word vectors for 157 languages.
\newblock In {\em Proceedings of the Eleventh International Conference on
  Language Resources and Evaluation (LREC 2018)}, Miyazaki, Japan.

\bibitem[Harm and Seidenberg, 2004]{harm2004computing}
Harm, M.~W. and Seidenberg, M.~S. (2004).
\newblock Computing the meanings of words in reading: cooperative division of
  labor between visual and phonological processes.
\newblock {\em Psychological review}, 111(3):662.

\bibitem[Heitmeier and Baayen, 2020]{heitmeier2020simulating}
Heitmeier, M. and Baayen, R.~H. (2020).
\newblock {Simulating phonological and semantic impairment of English tense
  inflection with linear discriminative learning}.
\newblock {\em The Mental Lexicon}, 15(3):385--421.

\bibitem[Heitmeier et~al., 2021]{heitmeier2021modeling}
Heitmeier, M., Chuang, Y.-Y., and Baayen, R.~H. (2021).
\newblock Modeling morphology with linear discriminative learning:
  considerations and design choices.
\newblock {\em Frontiers in Psychology}, 12.

\bibitem[Heitmeier et~al., 2023a]{heitmeier2023trial}
Heitmeier, M., Chuang, Y.-Y., and Baayen, R.~H. (2023a).
\newblock How trial-to-trial learning shapes mappings in the mental lexicon:
  Modelling lexical decision with linear discriminative learning.
\newblock {\em Cognitive Psychology}, 146:101598.

\bibitem[Heitmeier et~al., 2023b]{heitmeier2023linear}
Heitmeier, M., Chuang, Y.-Y., Luo, X., and Baayen, H. (2023b).
\newblock {Linear Discriminative Learning: Theory and implementation in the
  julia package JudiLing}.
\newblock {\em Manuscript, University of Tübingen}.

\bibitem[Ho, 1976]{ho1976acoustic}
Ho, A.~T. (1976).
\newblock The acoustic variation of {M}andarin tones.
\newblock {\em Phonetica}, 33(5):353--367.

\bibitem[Hollis, 2020]{hollis2020delineating}
Hollis, G. (2020).
\newblock Delineating linguistic contexts, and the validity of context
  diversity as a measure of a word's contextual variability.
\newblock {\em Journal of Memory and Language}, 114:104146.

\bibitem[Howie, 1974]{howie1974domain}
Howie, J.~M. (1974).
\newblock On the domain of tone in mandarin.
\newblock {\em Phonetica}, 30(3):129--148.

\bibitem[Jacobs and Grainger, 1994]{jacobs1994models}
Jacobs, A.~M. and Grainger, J. (1994).
\newblock Models of visual word recognition: sampling the state of the art.
\newblock {\em Journal of Experimental Psychology: Human perception and
  performance}, 20(6):1311.

\bibitem[Kapatsinski, 2022]{kapatsinski2022logistic}
Kapatsinski, V. (2022).
\newblock The logistic perceptron accounts for rank frequency effects in
  lexical processing.
\newblock In Nixon, J., Tomaschek, F., and Baayen, R.~H., editors, {\em
  Proceedings of the Second International Conference on Error-Driven Learning
  in Language (EDLL 2022)}, pages 16--17. Tübingen.

\bibitem[Keuleers et~al., 2010]{keuleers2010practice}
Keuleers, E., Diependaele, K., and Brysbaert, M. (2010).
\newblock Practice effects in large-scale visual word recognition studies: A
  lexical decision study on 14,000 dutch mono-and disyllabic words and
  nonwords.
\newblock {\em Frontiers in psychology}, 1:174.

\bibitem[Keuleers et~al., 2012]{keuleers2012british}
Keuleers, E., Lacey, P., Rastle, K., and Brysbaert, M. (2012).
\newblock The british lexicon project: Lexical decision data for 28,730
  monosyllabic and disyllabic english words.
\newblock {\em Behavior research methods}, 44(1):287--304.

\bibitem[Kuperman and Van~Dyke, 2013]{kuperman2013reassessing}
Kuperman, V. and Van~Dyke, J.~A. (2013).
\newblock Reassessing word frequency as a determinant of word recognition for
  skilled and unskilled readers.
\newblock {\em Journal of Experimental Psychology: Human Perception and
  Performance}, 39(3):802.

\bibitem[Landauer et~al., 1998]{Landauer1998}
Landauer, T., Foltz, P., and Laham, D. (1998).
\newblock Introduction to latent semantic analysis.
\newblock {\em Discourse Processes}, 25:259--284.

\bibitem[Lee, 2007]{lee2007does}
Lee, C.-Y. (2007).
\newblock Does horse activate mother? processing lexical tone in form priming.
\newblock {\em Language and Speech}, 50(1):101--123.

\bibitem[Li et~al., 2007]{li2007dynamic}
Li, P., Zhao, X., and Mac~Whinney, B. (2007).
\newblock Dynamic self-organization and early lexical development in children.
\newblock {\em Cognitive science}, 31(4):581--612.

\bibitem[Luo, 2021]{luo2021judiling}
Luo, X. (2021).
\newblock {JudiLing: An implementation for Linear Discriminative Learning in
  JudiLing (unpublished Master's thesis)}.

\bibitem[MacWhinney, 2014]{macwhinney2014childes}
MacWhinney, B. (2014).
\newblock {\em The CHILDES project: Tools for analyzing talk, Volume II: The
  database}.
\newblock Psychology Press, Hove.

\bibitem[McClelland and Rumelhart, 1981]{mcclelland1981interactive}
McClelland, J.~L. and Rumelhart, D.~E. (1981).
\newblock An interactive activation model of context effects in letter
  perception: I. an account of basic findings.
\newblock {\em Psychological review}, 88(5):375.

\bibitem[McClelland and Rumelhart, 1989]{mcclelland1989explorations}
McClelland, J.~L. and Rumelhart, D.~E. (1989).
\newblock {\em Explorations in parallel distributed processing: A handbook of
  models, programs, and exercises}.
\newblock MIT press, Cambridge, MA.

\bibitem[Milin et~al., 2020]{milin2020keeping}
Milin, P., Madabushi, H.~T., Croucher, M., and Divjak, D. (2020).
\newblock Keeping it simple: Implementation and performance of the
  proto-principle of adaptation and learning in the language sciences.
\newblock {\em arXiv preprint arXiv:2003.03813}.

\bibitem[Miwa and Baayen, 2021]{miwa2021nonlinearities}
Miwa, K. and Baayen, H. (2021).
\newblock Nonlinearities in bilingual visual word recognition: An introduction
  to generalized additive modeling.
\newblock {\em Bilingualism: Language and Cognition}, 24(5):825--832.

\bibitem[Morton, 1969]{morton1969interaction}
Morton, J. (1969).
\newblock Interaction of information in word recognition.
\newblock {\em Psychological review}, 76(2):165.

\bibitem[Morton, 1979a]{morton1979facilitation}
Morton, J. (1979a).
\newblock Facilitation in word recognition: Experiments causing change in the
  logogen model.
\newblock {\em Processing of visible language}, 13:259--268.

\bibitem[Morton, 1979b]{morton1979word}
Morton, J. (1979b).
\newblock Word recognition.
\newblock {\em Psycholinguistics: Series 2. Structures and processes}, pages
  107--156.

\bibitem[Murray and Forster, 2004]{murray2004serial}
Murray, W.~S. and Forster, K.~I. (2004).
\newblock Serial mechanisms in lexical access: the rank hypothesis.
\newblock {\em Psychological Review}, 111(3):721.

\bibitem[Nieder et~al., 2023]{nieder2023discriminative}
Nieder, J., Chuang, Y.-Y., van~de Vijver, R., and Baayen, H. (2023).
\newblock A discriminative lexicon approach to word comprehension, production,
  and processing: Maltese plurals.
\newblock {\em Language}, 99(2):242--274.

\bibitem[Norris, 2006]{norris2006bayesian}
Norris, D. (2006).
\newblock The bayesian reader: explaining word recognition as an optimal
  bayesian decision process.
\newblock {\em Psychological review}, 113(2):327.

\bibitem[Norris, 2013]{norris2013models}
Norris, D. (2013).
\newblock Models of visual word recognition.
\newblock {\em Trends in cognitive sciences}, 17(10):517--524.

\bibitem[Nusbaum, 1985]{Nusbaum:85}
Nusbaum, H.~C. (1985).
\newblock A stochastic account of the relationship between lexical density and
  word frequency.
\newblock Technical report, Indiana University.
\newblock Research on Speech Perception, Progress Report \#11.

\bibitem[Pham and Baayen, 2015]{pham2015vietnamese}
Pham, H. and Baayen, H. (2015).
\newblock Vietnamese compounds show an anti-frequency effect in visual lexical
  decision.
\newblock {\em Language, Cognition and Neuroscience}, 30(9):1077--1095.

\bibitem[{R Core Team}, 2020]{rcoreteam2020r}
{R Core Team} (2020).
\newblock {\em R: A Language and Environment for Statistical Computing}.
\newblock R Foundation for Statistical Computing, Vienna, Austria.

\bibitem[Ratcliff et~al., 2004]{ratcliff2004diffusion}
Ratcliff, R., Gomez, P., and McKoon, G. (2004).
\newblock A diffusion model account of the lexical decision task.
\newblock {\em Psychological review}, 111(1):159.

\bibitem[Reicher, 1969]{reicher1969perceptual}
Reicher, G.~M. (1969).
\newblock Perceptual recognition as a function of meaningfulness of stimulus
  material.
\newblock {\em Journal of experimental psychology}, 81(2):275.

\bibitem[Rescorla, 1967]{rescorla1967pavlovian}
Rescorla, R.~A. (1967).
\newblock Pavlovian conditioning and its proper control procedures.
\newblock {\em Psychological review}, 74(1):71.

\bibitem[Rescorla and Wagner, 1972]{Rescorla1972}
Rescorla, R.~A. and Wagner, A.~R. (1972).
\newblock {\em Classical conditioning II: current research and theory}, chapter
  A theory of Pavlovian conitioning: variations in the effectiveness of
  reinforcement and nonreinforcement, pages 64--99.
\newblock Appleton-Century-Crofts, New York.

\bibitem[Rubenstein et~al., 1970]{rubenstein1970homographic}
Rubenstein, H., Garfield, L., and Millikan, J.~A. (1970).
\newblock Homographic entries in the internal lexicon.
\newblock {\em Journal of verbal learning and verbal behavior}, 9(5):487--494.

\bibitem[Rubenstein et~al., 1971]{rubenstein1971homographic}
Rubenstein, H., Lewis, S.~S., and Rubenstein, M.~A. (1971).
\newblock Homographic entries in the internal lexicon: Effects of systematicity
  and relative frequency of meanings.
\newblock {\em Journal of verbal learning and verbal behavior}, 10(1):57--62.

\bibitem[Rumelhart et~al., 1986]{rumelhart1986backprop}
Rumelhart, D.~E., Hinton, G.~E., and Williams, R.~J. (1986).
\newblock Learning representations by back-propagating errors.
\newblock {\em Nature}, 323(6088):533--536.

\bibitem[Rumelhart and McClelland, 1982]{rumelhart1982interactive}
Rumelhart, D.~E. and McClelland, J.~L. (1982).
\newblock An interactive activation model of context effects in letter
  perception: Ii. the contextual enhancement effect and some tests and
  extensions of the model.
\newblock {\em Psychological review}, 89(1):60.

\bibitem[Sanchez et~al., 2019]{sanchez2019childes}
Sanchez, A., Meylan, S.~C., Braginsky, M., MacDonald, K.~E., Yurovsky, D., and
  Frank, M.~C. (2019).
\newblock childes-db: A flexible and reproducible interface to the child
  language data exchange system.
\newblock {\em Behavior research methods}, 51(4):1928--1941.

\bibitem[Schmitz et~al., 2021]{schmitz2021durational}
Schmitz, D., Plag, I., Baer-Henney, D., and Stein, S.~D. (2021).
\newblock Durational differences of word-final/s/emerge from the lexicon:
  Modelling morpho-phonetic effects in pseudowords with linear discriminative
  learning.
\newblock {\em Frontiers in Psychology}, 12:2983.

\bibitem[Seidenberg and McClelland, 1989]{seidenberg1989distributed}
Seidenberg, M.~S. and McClelland, J.~L. (1989).
\newblock A distributed, developmental model of word recognition and naming.
\newblock {\em Psychological review}, 96(4):523.

\bibitem[Shafaei-Bajestan et~al., 2023]{shafaei2021ldl}
Shafaei-Bajestan, E., Moradipour-Tari, M., Uhrig, P., and Baayen, R.~H. (2023).
\newblock Ldl-auris: a computational model, grounded in error-driven learning,
  for the comprehension of single spoken words.
\newblock {\em Language, Cognition and Neuroscience}, 38(4):509--536.

\bibitem[Shmueli, 2010]{shmueli2010explain}
Shmueli, G. (2010).
\newblock To explain or to predict?
\newblock {\em Statistical Science}, pages 289--310.

\bibitem[Stein and Plag, 2021]{stein2021morpho}
Stein, S.~D. and Plag, I. (2021).
\newblock Morpho-phonetic effects in speech production: Modeling the acoustic
  duration of english derived words with linear discriminative learning.
\newblock {\em Frontiers in Psychology}, 12.

\bibitem[Sun et~al., 2018]{sun2018chinese}
Sun, C.~C., Hendrix, P., Ma, J., and Baayen, R.~H. (2018).
\newblock Chinese lexical database (cld) a large-scale lexical database for
  simplified mandarin chinese.
\newblock {\em Behavior Research Methods}, 50:2606--2629.

\bibitem[Tomaschek et~al., 2018]{Tomaschek:Tucker:Baayen:2018}
Tomaschek, F., Tucker, B., and Baayen, R.~H. (2018).
\newblock Practice makes perfect: The consequences of lexical proficiency for
  articulation.
\newblock {\em Linguistic Vanguard}, 4:1--13.

\bibitem[van~de Vijver et~al., 2023]{van2023comprehension}
van~de Vijver, R., Uwambayinema, E., and Chuang, Y.-Y. (2023).
\newblock Comprehension and production of kinyarwanda verbs in the
  discriminative lexicon.
\newblock {\em Linguistics}.

\bibitem[van~der Velde et~al., 2022]{vanRijn2022capturing}
van~der Velde, M., Sense, F., Borst, J.~P., van Maanen, L., and Van~Rijn, H.
  (2022).
\newblock Capturing dynamic performance in a cognitive model: Estimating act-r
  memory parameters with the linear ballistic accumulator.
\newblock {\em Topics in Cognitive Science}, 14(4):889--903.

\bibitem[Widrow and Hoff, 1960]{Widrow1960}
Widrow, B. and Hoff, M. (1960).
\newblock Adaptive switching circuits.
\newblock {\em 1960 WESCON Convention Record Part IV}.

\bibitem[Wood, 2011]{wood2011gam}
Wood, S.~N. (2011).
\newblock Fast stable restricted maximum likelihood and marginal likelihood
  estimation of semiparametric generalized linear models.
\newblock {\em Journal of the Royal Statistical Society (B)}, 73(1):3--36.

\bibitem[Wood, 2017]{Wood:2017}
Wood, S.~N. (2017).
\newblock {\em Generalized additive models: an introduction with R}.
\newblock CRC press.

\bibitem[Xu and Wang, 2001]{xu2001pitch}
Xu, Y. and Wang, Q.~E. (2001).
\newblock Pitch targets and their realization: Evidence from {Mandarin
  Chinese}.
\newblock {\em Speech communication}, 33(4):319--337.

\bibitem[Yamada et~al., 2020]{yamada2020wikipedia2vec}
Yamada, I., Asai, A., Sakuma, J., Shindo, H., Takeda, H., Takefuji, Y., and
  Matsumoto, Y. (2020).
\newblock {W}ikipedia2{V}ec: An efficient toolkit for learning and visualizing
  the embeddings of words and entities from {W}ikipedia.
\newblock In {\em Proceedings of the 2020 Conference on Empirical Methods in
  Natural Language Processing: System Demonstrations}, pages 23--30, online.
  Association for Computational Linguistics.

\end{thebibliography}

\end{document}